\documentclass[10pt,twocolumn,twoside]{IEEEtran}
\usepackage{diagbox}
\usepackage{bm}
\usepackage{hyperref}
\usepackage{mathtools}
\usepackage{amsfonts,amssymb}
\usepackage{threeparttable}
\usepackage{makecell}
\usepackage{multirow}
\usepackage{bbding}
\usepackage{booktabs}                 
\usepackage{algorithmic}
\usepackage[linesnumbered,ruled,vlined]{algorithm2e}

\usepackage{graphicx}
\usepackage{amsmath}
\usepackage{textcomp}
\usepackage{xcolor}
\def\BibTeX{{\rm B\kern-.05em{\sc i\kern-.025em b}\kern-.08em
    T\kern-.1667em\lower.7ex\hbox{E}\kern-.125emX}}

\usepackage{adjustbox}
\usepackage{multirow}
\usepackage{amssymb,mathtools}

\usepackage{xurl}

\ifCLASSINFOpdf
\else
\fi
\makeatletter
\def\footnoterule{\kern-3\p@
  \hrule \@height 1pt \@width 3.5in \kern 2.6\p@} 
\makeatother

\ExplSyntaxOn

\cs_new_eq:NN \guillaume_footnoterule: \footnoterule
\cs_set_protected:Npn \footnoterule
 {
  \bool_set_true:N \l_guillaume_footnoterule_bool
  \seq_map_function:NN \g_guillaume_footnoterule_pages_seq \guillaume_test:n
  \bool_if:NT \l_guillaume_footnoterule_bool { \guillaume_footnoterule: }
 }
\cs_new_protected:Nn \guillaume_test:n
 {
  \int_compare:nT { #1 = \value{page} }
   {
    \seq_map_break:n { \bool_set_false:N \l_guillaume_footnoterule_bool }
   }
 }
\seq_new:N \g_guillaume_footnoterule_pages_seq
\NewDocumentCommand{\nofootnoterule}{m}
 {
  \seq_gset_from_clist:Nn \g_guillaume_footnoterule_pages_seq { #1 }
 }
\ExplSyntaxOff

\nofootnoterule{1}

\begin{document}
\bstctlcite{IEEEexample:BSTcontrol}

\title{A Novel DDPM-based Ensemble Approach for Energy Theft Detection in Smart Grids}

%
%
%

\author{Xun Yuan,
        Yang Yang,
        Asif Iqbal,\\
        Prosanta Gope,~\IEEEmembership{Senior Member,~IEEE}
        and~Biplab~Sikdar,~\IEEEmembership{Senior Member,~IEEE}
\thanks{Xun Yuan, Yang Yang, Asif Iqbal, and Biplab Sikdar are with the Department of Electrical and Computer Engineering, College of Design and Engineering, National University of Singapore, Singapore. (E-mail: e0919068@u.nus.edu, y.yang@u.nus.edu, aiqbal@nus.edu.sg, bsikdar@nus.edu.sg).}
\thanks{Prosanta Gope is with the Department of Computer Science, University of Sheffield, United Kingdom. (E-mail: p.gope@sheffield.ac.uk).}

}

%
%

\markboth{Preprint}%
{Yuan \MakeLowercase{\textit{et al.}}: DDPM based approach for ETD in Smart Grid}
%



\maketitle

\begin{abstract}

Energy theft, characterized by manipulating energy consumption readings to reduce payments, poses a dual threat—causing financial losses for grid operators and undermining the performance of smart grids. Effective Energy Theft Detection (ETD) methods become crucial in mitigating these risks by identifying such fraudulent activities in their early stages.
However, the majority of current ETD methods rely on supervised learning, which is hindered by the difficulty of labelling data and the risk of overfitting known attacks.
To address these challenges, several unsupervised ETD methods have been proposed, focusing on learning the normal patterns from honest users, specifically the reconstruction of input. However, our investigation reveals a limitation in current unsupervised ETD methods, as they can only detect anomalous behaviours in users exhibiting regular patterns. Users with high-variance behaviours pose a challenge to these methods.
In response, this paper introduces a Denoising Diffusion Probabilistic Model (DDPM)-based ETD approach. This innovative approach demonstrates impressive ETD performance on high-variance smart grid data by incorporating additional attributes correlated with energy consumption. The proposed methods improve the average ETD performance on high-variance smart grid data from below 0.5 to over 0.9 w.r.t. AUC.
On the other hand, our experimental findings indicate that while the state-of-the-art ETD method(s) based on reconstruction error can identify ETD attacks for the majority of users, they prove ineffective in detecting attacks for certain users.  To address this, we propose a novel ensemble approach that considers both reconstruction error and forecasting error, enhancing the robustness of the ETD methodology. The proposed ensemble method improves the average ETD performance on the stealthiest attacks from nearly 0 to 0.5 w.r.t. 5\%-TPR.

\end{abstract}


%
\IEEEpeerreviewmaketitle

\section{Introduction}

\IEEEPARstart{S}{mart} grids represent an advanced power infrastructure integrating digital communication technology, smart sensors, artificial intelligence, and big data analytics with traditional power grid systems. This amalgamation significantly elevates conventional grids' efficiency, reliability, and security. A smart grid effectively addresses inherent limitations in traditional grid systems by incorporating intelligent optimisation techniques, such as demand-response management. Furthermore, deploying intelligent technologies within the smart grid enables the implementation of anomaly detection methods to identify and mitigate energy thefts, thereby fortifying security and performance.

Energy thefts, encompassing diverse methods aimed at reducing electricity payments or obtaining unauthorized financial benefits from the smart grid, pose a considerable threat and result in substantial financial consequences for energy companies. Additionally, energy theft can disrupt the demand-response capabilities of the grid, impacting its ability to accurately assess real power consumption and posing potential risks. The security of the smart grid is susceptible to compromise in the presence of an imbalance between generation and demand. A recent study quantifies the monetary losses from energy theft in the UK and the US, reaching up to US\$ 6 billion \cite{Yang_ref1}. Furthermore, 15 power outages in the US in 2017 were attributed to electricity theft \cite{Yang_ref2}, emphasizing the urgent need to address this issue.
Therefore, detecting and preventing energy theft is of utmost importance in the smart grid. Energy theft detection (ETD) plays a critical role in defending against such threats, enabling smart grid companies to predict electricity demand more accurately. This, in turn, maximizes the utilization of limited resources \cite{Yang_ref4}, resulting in cost savings for grid companies and promoting environmentally friendly energy consumption \cite{Yang_ref3}.

Many existing energy theft detection methods rely on supervised learning, which is prone to overfitting and constrained by the difficulty of well-labelled datasets, i.e., labelling a dataset for energy theft detection is time-consuming and needs effort from domain experts \cite{takiddin2020robust}. To address the limitations of supervised ETD methods, contemporary unsupervised ETD techniques leverage data reconstruction to learn normal patterns of smart grid data and identify energy theft behaviours through reconstruction errors. 
Nevertheless, Reconstruction Error-based Methods (REMs) exhibit limitations in detecting anomalies for specific users. This study discovers that an ensemble approach considering both reconstruction error and forecasting error can address this constraint. 
On the other hand, current unsupervised ETD methods are ineffective for high-variance smart grid data. For instance, certain residents may adopt irregular lifestyles, resulting in daily fluctuations in energy consumption. Typically, energy consumption forecasting methods \cite{kumar2018energy, wei2019conventional} encounter challenges in accurately forecasting future energy consumption when confronted with high-variance smart grid data.
To resolve the above issues, this work introduces a Denoising Diffusion Probabilistic Model (DDPM)-based approach for robust energy theft detection. In summary, the DDPM empowers the proposed method to effectively handle high-variance data. The ensemble approach, which takes into account both reconstruction and forecasting errors, further enhances the overall performance of the method across all user profiles.


\subsection{Related Work and Motivation} \label{sec:related work}
\begin{table*}[htbp]
\renewcommand\arraystretch{1.}
    \caption{Summary of the Related Works}
    \label{tab:related papers}
    \centering
    \begin{tabular}{c c c c c c c c}
        \hline
        \multirow{2}{*}{\makebox[0.15\textwidth]{\textbf{Scheme}}} &\multirow{2}{*}{\makebox[0.15\textwidth]{\textbf{Proposed Approach}}}  & \multicolumn{2}{c}{\makebox[0.16\textwidth]{\textbf{Supported Data Types}}} & \multicolumn{4}{c}{\makebox[0.32\textwidth]{\textbf{ETD Mechanisms Applied}}}\\
        \cmidrule(lr){3-4}\cmidrule(lr){5-8}
        {} & {} & \makebox[0.08\textwidth]{\textbf{H-V}} & \makebox[0.08\textwidth]{\textbf{Regular}} & \makebox[0.08\textwidth]{\textbf{UL}} & \makebox[0.08\textwidth]{\textbf{REM}} & \makebox[0.08\textwidth]{\textbf{FEM}} & \makebox[0.08\textwidth]{\textbf{EM}}\\
        \hline
        Nabil et al. \cite{ref15} & RNN & \XSolidBrush & \Checkmark & \XSolidBrush & \XSolidBrush & \XSolidBrush & \XSolidBrush \\
        \hline
        Ullah et al. \cite{ref18} & GRU \& CNN & \XSolidBrush & \Checkmark & \XSolidBrush & \XSolidBrush & \XSolidBrush & \XSolidBrush\\
        \hline        
        Gao et al. \cite{ref13} & ConvLSTM & \XSolidBrush & \Checkmark & \XSolidBrush & \XSolidBrush & \XSolidBrush & \XSolidBrush\\
        \hline
        Alromih et al. \cite{ref9} & FC & \XSolidBrush & \Checkmark & \Checkmark & \Checkmark & \XSolidBrush & \XSolidBrush\\
        \hline
        Takiddin et al. \cite{ref10} & VAE & \XSolidBrush & \Checkmark & \Checkmark & \Checkmark & \XSolidBrush & \XSolidBrush\\
        \hline
        Ours & DDPM & \Checkmark & \Checkmark & \Checkmark & \Checkmark & \Checkmark & \Checkmark\\
        \hline
    \end{tabular}
    \begin{tablenotes}
        \footnotesize
        \item \textbf{H-V}: High-variance; \textbf{ETD}: Energy theft detection; \textbf{UL}: Unsupervised learning; \textbf{REM}: Reconstruction error-based method;
        \textbf{FEM}: Forecasting error-based method; \textbf{EM}: Ensemble method.
    \end{tablenotes}
\end{table*}

In this section, we provide a literature review of contemporary energy theft detection methods and identify research gaps. Subsequently, we articulate the motivation for this paper based on the identified research gaps.

\subsubsection{\textbf{Supervised ETD methods}} 

Most existing ETD models \cite{ref13,ref15,ref18,ref19,ref16} leverage \emph{supervised} learning. For instance, in \cite{ref15}, the authors assert the superior performance of Recurrent Neural Networks (RNN) over shallow machine learning approaches, incorporating synthetic attacks for model training. In \cite{ref18, ref19}, CNN-RNN-based models are proposed, with the application of Synthetic Minority Over-sampling Technique (SMOTE) \cite{ref40} to generate attack samples. The work in \cite{ref13} introduces a convLSTM model, claiming its superiority over CNN-RNN methods, and adopts the borderline-SMOTE \cite{ref14} sampling technique to generate more realistic energy theft data than SMOTE. Additionally, in \cite{ref16}, the authors present an evolutionary hyper-parameter tuning method for deep RNN models, utilizing an Adaptive Synthetic Sampling Approach (ADASYN) \cite{ref20} to address dataset imbalance.

While the majority of ETD methods rely on supervised learning, these approaches encounter practical challenges due to inherent flaws of supervised learning: (1) labelling energy theft data is time-consuming and needs effort from smart grids experts; (2) energy-theft data is typically scarce, leading to imbalanced datasets which significantly impacts the performance of supervised learning methods; (3) overfitting of anomaly samples in the training data is a common issue with supervised learning methods; (4) these methods are less effective at detecting unseen attacks. Although some prior studies address issues (1) and (2) by generating synthetic attack data, they often overlook or fail to address shortcomings related to (3) and (4).

\subsubsection{\textbf{Unsupervised ETD methods}} 

Research on unsupervised ETD methods is limited, and current unsupervised ETD methods primarily rely on reconstruction error. These methods involve reconstructing the input and computing the reconstruction error, denoted as the distance between the reconstruction result and the corresponding input. Such methods, termed Reconstruction Error-based Methods (REMs), identify energy thefts if the reconstruction error exceeds a predefined threshold. In \cite{ref9}, the authors use a Fully Connected (FC) neural network for reconstruction, and energy thefts are identified based on the reconstruction error. In \cite{ref10}, an LSTM-based Variational AutoEncoder (VAE) \cite{ref29} is employed for reconstruction. However, we observe that REMs are ineffective for certain users even with regular behaviour. To address this limitation, we introduce the Forecasting Error-based Method (FEM), which is beneficial in detecting energy thefts that may go undetected by REMs. Specifically, FEMs predict future energy consumption and calculate the forecasting error according to the distance between the forecasting result and the ground truth. Like REMs, energy thefts are identified if the forecasting error surpasses a predetermined threshold.

While the experimental results in \cite{ref9} indicate the suitability of their method for high-variance smart grid data, it is noteworthy that energy theft attacks were exclusively applied to the `energy consumption'. We discovered that the curve for the `current' is very similar to that of `energy consumption', suggesting potential information leakage. In this study, we address this concern by conducting energy theft attacks on both `energy consumption' and `current'. Contrary to the findings in \cite{ref9} and \cite{ref10}, our experiments reveal that their proposed methods fail to detect most energy theft attacks for high-variance smart grid data.
To fortify our conclusions, we adapt the LSTM-based multi-sensor anomaly detection method \cite{ref37} to an REM for ETD. Additionally, we design an LSTM-based FEM for ETD by modifying the LSTM-based energy load forecasting method proposed in \cite{ref38}. Unfortunately, both LSTM-based REM and FEM are ineffective in detecting energy thefts in high-variance smart grid data.

\subsubsection{\textbf{Motivations}}

Energy theft detection is critical for safeguarding smart grids against energy theft attacks. Despite numerous solutions proposed in existing literature for ETD, they grapple with various limitations. Primarily, a majority of proposed ETD solutions rely on supervised learning for model training, making them susceptible to imbalanced data and overfitting issues. Secondly, there is a lack of research on unsupervised learning approaches for ETD, with most falling under the REM category, presenting ineffectiveness in detecting energy theft in certain users. Thirdly, the potential advantages of employing forecasting error for energy theft detection remain unexplored. Most importantly, current unsupervised ETD methods cannot identify energy theft attacks in high-variance smart grid data. Lastly, although DDPM has demonstrated success in image anomaly detection \cite{ref31,ref32}, their application to the ETD problem is yet to be studied. Our objective is to assess the potential of DDPM in addressing the ETD problem and bridging the identified research gaps.

\subsection{Contributions} \label{sec:contributions}
In response to the above-mentioned research gaps, the contributions of this paper can be summarized as follows:

\begin{itemize}
\item[$\bullet$] We propose a DDPM-based unsupervised ensemble approach for energy theft detection, termed as \textit{ETDddpm}, which considers both reconstruction and forecasting errors. Remarkably, this is the \emph{first} work to show how DDPM can address the ETD problem.

\item[$\bullet$] The proposed ensemble approach shows consistently impressive ETD performance for all users, while single REM and FEM show limitations on some users. 
To the best of our knowledge, this is the \emph{first} work to show how FEM deals with the ETD problem and this is the \emph{first} work combining REM and FEM for the ETD problem.

\item[$\bullet$]The proposed \textit{ETDddpm} delivers impressive ETD performance on high-variance smart grid data, where current ETD methods fail to work.

\item[$\bullet$] This paper introduces a unified learning objective for the training of \textit{ETDddpm} to optimise the model's capabilities in both reconstruction and forecasting. By integrating these dual objectives, we exploit the inherent interdependencies and shared information between the tasks of reconstruction and forecasting, thereby augmenting the model's overall performance.

\end{itemize}

The remainder of this paper is organized as follows. Section \ref{sec:2} introduces the preliminaries for this paper. In Section \ref{sec:3}, we describe our proposed method in detail, including the model architecture, training process, and inference process.
In Section \ref{sec:4}, we evaluate the proposed method on two datasets: a real-world dataset and a synthetic smart grid dataset. Lastly, we present the conclusion in Section \ref{sec:5}.

\section{Preliminaries} \label{sec:2}

This section first introduces how the ETD problem is transformed into an optimization problem that minimizes the reconstruction and forecasting error. Then, we specify the adversary model, considering seven attack scenarios. After that, we describe our system model for ETD. Finally, we introduce the mechanism of DDPM, which is the underlying foundation of the proposed \textit{ETDddpm}.

\subsection{Problem Formulation} \label{problem statement}

In this paper, we consider three assumptions for transforming the ETD problem into an optimization problem that minimizes the reconstruction and forecasting error.

\textbf{Assumption 1:} In this assumption for REM, we assume that anomalies cannot be effectively reconstructed with a minor error since information is lost in the mapping from the input space to the latent space. 

\textbf{Assumption 2:} In this assumption for FEM, we assume that anomalous values cannot be correctly predicted as normal ones. 

\textbf{Assumption 3:} The training dataset exclusively consists of honest data (the dataset without energy theft).

With the above assumptions, we can train deep learning models with the training dataset for reconstruction and forecasting, and energy thefts can be detected when the reconstruction error or forecasting error of the models exceeds a predefined threshold. In order to formulate the optimization problem, we first show how we represent the smart meter data. Smart grid data can be represented by time series with look-back window $L$ as $\bm x_{1:L}=(\bm x_1,\bm x_2,\cdots,\bm x_L)$ where each $\bm x_l$ at time step $l$ is a vector of dimension $M$ for multivariate data or a real number for univariate data. Then, the REM and FEM for the ETD problem can be defined as follows.

\textbf{REM} reconstructs the input sequence $\bm x_{1:L}$ into an output sequence $\hat{\bm x}_{1:L}$, and then computes the mean absolute error (MAE) between the input and reconstruction sequences as anomaly score,
\begin{equation}
    \label{error_R}
    \delta_R={\rm mean}(|\hat{\bm x}_{1:L}-\bm x_{1:L}|).
\end{equation}
If $\delta_R$ is greater than a manually set threshold $th_R$, we classify the input as an anomaly. 

\textbf{FEM} forecasts the future sequence of the input for the next $T$ time steps, i.e., $\hat{\bm x}_{L+1:L+T}$, and then computes the MAE between the real future data, $\bm x_{L+1:L+T}$, and the forecasting sequence as anomaly score,
\begin{equation}
    \label{eq:error_F}
    \delta_F={\rm mean}(|\hat{\bm x}_{L+1:L+T}-\bm x_{L+1:L+T}|).
\end{equation}
Similarly, if $\delta_F$ is greater than a manually set threshold $th_F$, we classify the input as an anomaly. 

According to the assumptions, the optimization objectives of REM and FEM for the ETD problem should be minimizing \eqref{error_R} and \eqref{eq:error_F}, respectively. For the ensemble method proposed in this study, the reconstruction and forecasting errors should be minimized simultaneously. Thus, the optimization problem of the ensemble method can be expressed as,
\begin{equation}
    \label{eq:unified problem}
    \mathrm{P:}\quad \min_\theta(\delta_R+\gamma\delta_F)
\end{equation}
where $\theta$ denotes the model's parameters and $\gamma$ is a balancing coefficient. With this objective and the above-mentioned assumptions, the model can reconstruct and forecast the normal data well and the reconstruction and forecasting errors become larger when energy theft attacks occur.

\subsection{Adversary Model} \label{sec:attack methods}
Energy theft attacks can be accomplished by manipulating readings of energy consumption. In our adversary model, we consider that a malicious user can change his/her energy consumption readings to launch a successful energy theft attack. This includes seven attack scenarios that have been adopted from \cite{ref9,ref10}, i.e., (1) fixed reduction, (2) partial reduction, (3) random partial reduction, (4) random average consumption, (5) average consumption, (6) reverse, and (7) selective by-pass. 

Under a ‘fixed reduction’ attack, an adversary may attempt to subtract the normal data $\bm x_{1:L}$ with a fixed value,
\begin{equation}
    \label{eq:attack 1}
    f_1(\bm x_{1:L}) =\max(\bm x_{1:L} - \gamma_1  \mathbb{E}(\bm x),\bm 0),
\end{equation}
where $\mathbb{E}(\bm x)$ denotes the mean of the normal data and $\gamma_1$ is set to 0.2 while it's set to 0.4 in \cite{ref9}.

Under a ‘partial reduction’ attack, we consider an adversary who multiplies the normal data by a fixed coefficient,
\begin{equation}
    \label{eq:attack 2}
    f_2(\bm x_{1:L}) =\gamma_2\bm x_{1:L},
\end{equation}
where $\gamma_2$ is set to 0.8 while it's randomly sampled from [0.1,0.8] in \cite{ref10}. Thus, our adversary model is stealthier.

Under a `random partial reduction’ attack, an adversary multiplies the normal data by a random coefficient,
\begin{equation}
    \label{eq:attack 3}
    f_3(\bm x_{1:L}) = \mathrm{rand}(\mathrm{min}=\gamma_{31},\mathrm{max}=\gamma_{32})\bm x_{1:L},
\end{equation}
where $\mathrm{rand(\cdot)}$ uniformly chooses a value from the range $[\gamma_{31},\gamma_{32}]$. $\gamma_{31}$ and $\gamma_{32}$ are set to 0.7 and 0.9, respectively, to obtain a similar mean value to the above ‘partial reduction’ attack. In \cite{ref10}, $\gamma_{31}$ and $\gamma_{32}$ are set to 0.1 and 0.8.

The ‘random average consumption' attack can be expressed as,
\begin{equation}
    \label{eq:attack 6}
    f_4(\bm x_{1:L}) = \mathrm{rand}(\mathrm{min}=\gamma_{31},\mathrm{max}=\gamma_{32})\mathbb{E}(\bm x).
\end{equation}
where $\gamma_{31}$ and $\gamma_{32}$ are the same as \eqref{eq:attack 3}. 

Under an ‘average consumption’ attack, an adversary reports the average consumption to the server,
\begin{equation}
    \label{eq:attack 4}
    f_5(\bm x_{1:L}) = \mathbb{E}(\bm x).
\end{equation}
Thus, the artificial data becomes a horizontal line.

Under a ‘reverse’ attack, an attacker reverses the original sequence every 24 hours. So for every $i \in \mathbb{N}$ and $24i$ less than the length of the whole sequence, we have
\begin{equation}
    \label{eq:attack 5}
    f_6(\bm x_{24(i-1)+1:24i}) = \mathrm{reverse}(\bm x_{24(i-1)+1:24i}).
\end{equation}

Under a 'selective bypass' attack, zero energy consumption is reported during an interval of time [$t_s, t_e$], and the true energy consumption is reported outside that interval. So, for all $i\in \{1,2,\cdots,L\}$ we have
\begin{equation}
    \label{eq:attack 7}
    f_7(\bm x_{i})=\left\{
    \begin{aligned}
             &0, &i\in[t_s, t_e] \\
             &\bm x_{i}&i\notin[t_s, t_e]
    \end{aligned}.
\right.
\end{equation}
We set $t_e-t_s=6$ in this paper.

\textbf{Remark:} The parameters of the above attack methods are more challenging than those in \cite{ref9, ref10} since our proposed methods and baseline methods achieve almost perfect performance when using the parameters described in \cite{ref9, ref10}.
\begin{figure}[htbp]
\center{\includegraphics[width=\linewidth, scale=1.2]{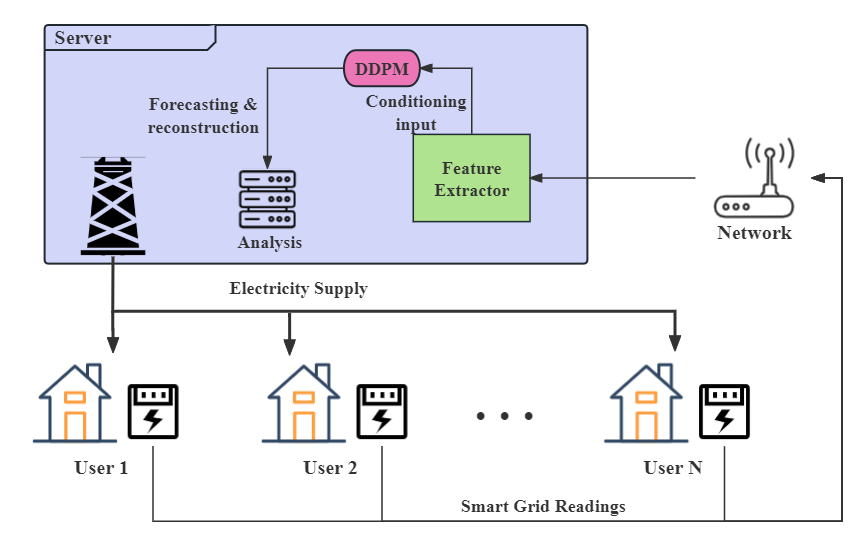}}
\caption{System model.}
\label{fig:system-model}
\end{figure}

\subsection{System Model}

The system model for energy theft detection in smart grids is depicted in Fig. \ref{fig:system-model}. The proposed \textit{ETDddpm} is comprised of two modules: the feature extractor and the DDPM module. The system is comprised of two key components: a group of users equipped with smart meters and a server responsible for electricity supply and energy theft detection. 
The users transmit their electricity consumption readings to the server through the Internet. In the server, the feature extractor first extracts features from the raw readings, and the features serve as the conditioning input for the denoising network of the DDPM module. Subsequently, the DDPM generates both reconstruction and forecasting sequences. Finally, the Analysis module calculates reconstruction and forecasting errors to identify energy theft behaviours.

\subsection{Denoising Diffusion Probabilistic Model (DDPM)} \label{preliminaries}
In this section, we provide the necessary preliminaries about the DDPM. For simplification, the notations in this section are different from those in Section \ref{sec:attack methods} and Section \ref{problem statement}, and the notations will be unified in Section \ref{LSTM-Diffusion Method}.

Formally, $\bm x^0 \sim q_{\mathcal{X}}(\bm x^0)$ denotes a vector from some input space $\mathcal{X} = \mathbb{R}^D$, $D=M$ for multivariate data and $D=1$ for univariate data.
The superscript represents the step of the diffusion process, e.g., 0 means the $0^{th}$ step.
$\bm x^0$ represents the ground truth of what we want to get from DDPM, i.e., reconstruction and forecasting sequences in this paper.
The output of DDPM, $p_{\theta}(\bm x^0)$ where $\theta$ denotes the model parameters, is a probability density function (PDF) that aims to approximate the real distribution of $\bm x^0$, $q_{\mathcal{X}}(\bm x^0)$. This optimization problem can be expressed as:
\begin{equation}
    \label{ddpm objective 1}
    \max \mathcal{L} \coloneqq \mathbb{E}_{q_{\mathcal{X}}(\bm x^0)}[-\log p_{\theta}(\bm x^0)].
\end{equation}

DDPM has two separate processes, i.e., diffusion and denoising. In the diffusion process, a fixed set of increasing variance parameters, $\beta\coloneqq\{\beta_1,\cdots,\beta_N\}$, is used to add Gaussian noise to $\bm x^{n-1}$ and obtain $\bm x^{n}$. The following equation does this:
\begin{equation}
    \label{diffusion 1}
    q(\bm x^n|\bm x^{n-1}) \coloneqq \mathcal{N}(\bm x^n; \sqrt{1-\beta_n}\bm x^{n-1}, \beta_n \mathrm{\mathbf{I}}),
\end{equation}
where $\mathrm{\mathbf{I}}$ is the identity matrix.
After adding noise by \eqref{diffusion 1} for $N$ steps, we get the diffusion sequences $\bm x^{0:N}$, where $p(\bm x^N) \simeq \mathcal{N}(\bm x^N; \mathrm{\mathbf{0}},\mathrm{\mathbf{I}})$. Reparameterizing \cite{falorsi2019reparameterizing} is a common strategy in deep learning. With the help of reparameterizing, $\bm x^n$ can be calculated in only one step for any given $n$:
\begin{equation}
    \label{eq:Reparameterizing}
    \bm x^n=\sqrt{\overline{\alpha}_n}\bm x^0+\sqrt{1-\overline{\alpha}_n}\bm\epsilon,
\end{equation}
where $\alpha_n\coloneqq 1-\beta_n$, $\overline{\alpha}_n\coloneqq \prod_{s=1}^{n}\alpha_s$, and $\bm\epsilon \sim \mathcal{N}(\mathrm{\mathbf{0}},\mathrm{\mathbf{I}})$.

The denoising process starts from $\bm x^N$, and denoises the data for $N$ steps to approximate the PDF of $\bm x^0$ using the following equation:
\begin{equation}
    \label{denoise 1}
    p_{\theta}(\bm x^{n-1}|\bm x^n)\coloneqq \mathcal{N}(\bm x^{n-1}; \mu_{\theta}(\bm x^n,n), \sigma_{\theta} \mathrm{\mathbf{I}}),
\end{equation}
where $\mu_{\theta}(\cdot)$ is a function that generates the mean value of the Gaussian distribution, $\sigma_{\theta}$ can be calculated by a function or a fixed number,  and $n$ denotes the diffusion step. 

As proved in \cite{ref3}, problem \eqref{ddpm objective 1} can be simplified as:
\begin{equation}
\begin{aligned}
    \label{ddpm objective simple}
    &\min \mathcal{L}_{simple}(\theta) \coloneqq  \\ &\mathbb{E}_{n,\bm x^0,\epsilon}\left[\left\Vert\frac{1}{\alpha_n}\left(\bm x^n-\frac{\beta_n}{\sqrt{1-\overline{\alpha}_n}}\epsilon \right)-\mu_{\theta}(\bm x^n,n)\right\Vert^2\right].
\end{aligned}
\end{equation}
Since $\bm x^n$, $\beta_n$, and $\overline{\alpha}_n$ in \eqref{ddpm objective simple} are known, we can use a deep learning model, $\bm\epsilon_{\theta}(\cdot)$, to approximate $\bm \epsilon$ instead of estimating $\mu_{\theta}(\bm x^n,n)$. Thus, problem \eqref{ddpm objective simple} can be rewritten as: 
\begin{equation}
\begin{aligned}
    \label{ddpm objective simple 2}
    \min \mathcal{L}_{simple}(\theta) \coloneqq  \mathbb{E}_{n,\bm x^0,\epsilon}[\Vert\bm \epsilon-\bm \epsilon_{\theta}(\bm x^n,n)\Vert^2].
\end{aligned}
\end{equation}
According to \cite{ref3}, optimizing \eqref{ddpm objective simple 2} obtains better performance than optimizing \eqref{ddpm objective simple}. Lastly, given $\bm x^n$ and $\bm \epsilon_{\theta}(\cdot)$, we can sample $\bm x^{n-1}$ by:
\begin{equation}
    \label{eq:denoising}
    \bm x^{n-1} = \frac{1}{\sqrt{\alpha}_n}\left(\bm x^n-\frac{\beta_n}{\sqrt{1-\overline{\alpha}_n}}\bm\epsilon(\bm x^n,n) \right) + \sigma_{\theta}\bm z,
\end{equation}
where $\bm z\sim \mathcal{N}(\mathrm{\mathbf{0}},\mathrm{\mathbf{I}})$. Computing \eqref{eq:denoising} for $N$ steps recurrently (changing $n$ from $N$ to 1), we can obtain $p_{\theta}(\bm x^0)$ from $p(\bm x^N) \simeq \mathcal{N}(\bm x^N; \mathrm{\mathbf{0}},\mathrm{\mathbf{I}})$, i.e., we can obtain our desired result from Gaussian noise.

\section{Proposed \textit{ETDddpm} Approach} \label{LSTM-Diffusion Method} \label{sec:3}

\begin{figure*}[htbp]
\centering
\includegraphics[width=\linewidth, scale=0.8]{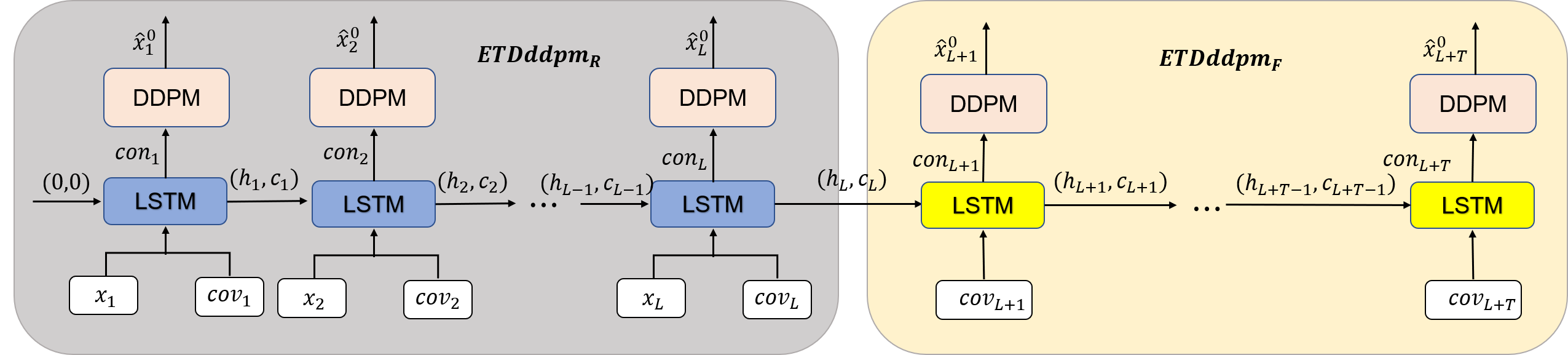}
\caption{The overall structure of proposed \textit{ETDddpm} where the left part is for reconstruction and the right part is for forecasting.}
\label{fig:diffusion comb}
\end{figure*}

As discussed in Section \ref{preliminaries}, DDPM can estimate the distribution of an observation, $\bm x_l^0$, from an energy consumption sequence, i.e., $\bm x_l^0 \in \{\bm x_1^0,\bm x_2^0,\cdots,\bm x_L^0\}$ and we sample the mean value of the distribution as the estimation of $\bm x_l^0$ (or $\bm x_l$), which is denoted by $\hat{\bm x}_l^0$. The subscript of $\bm x_l^0$ denotes the time step of smart grid data, and the superscript denotes the diffusion step. According to Section \ref{problem statement}, the objective of the ETD problem is to reconstruct the sequence $\bm x_{1:L}$ and forecast for the next $T$ time steps $\bm x_{L+1:L+T}$. 
In the proposed \textit{ETDddpm}, we have two sub-models, i.e., \textit{ETDddpm$_{R}$} and \textit{ETDddpm$_{F}$}, to produce the reconstruction sequence and the forecasting sequence, respectively. These sub-models are shown in Fig. \ref{fig:diffusion comb} and can be expressed as follows:
\begin{gather}
\label{objective reconstruction}
\hat{\bm x}_{1:L}^0,\bm h_L, \bm c_L = \mathrm{ETDddpm_{R}}({\bm x}_{1:L},\bm {cov}_{1:L}),\\
\label{objective prediction}
\hat{\bm x}_{L+1:L+T}^0 = \mathrm{ETDddpm_{F}}(\bm h_L, \bm c_L, \bm {cov}_{L+1:L+T}),
\end{gather}
where $\bm {cov}_l$ denotes the covariance of the observation at time step $l$. In this paper, the covariance contains temporal information like \cite{ref7}. Combining \textit{ETDddpm$_R$} and \textit{ETDddpm$_F$}, the proposed \textit{ETDddpm} can be expressed as: 
\begin{equation} 
    \label{eq:ETDddpm} 
    \bm{\hat{x}}_{1:L+T}^0 = \mathrm{ETDddpm}(\bm x_{1:L},\bm {cov}_{1:L+T}).
\end{equation}
In Fig. \ref{fig:diffusion comb}, blocks with the same colour are the same module, and blocks with different colours are different modules. 
\begin{figure}[htbp]
\centering
\includegraphics[width=\linewidth, scale=0.7]{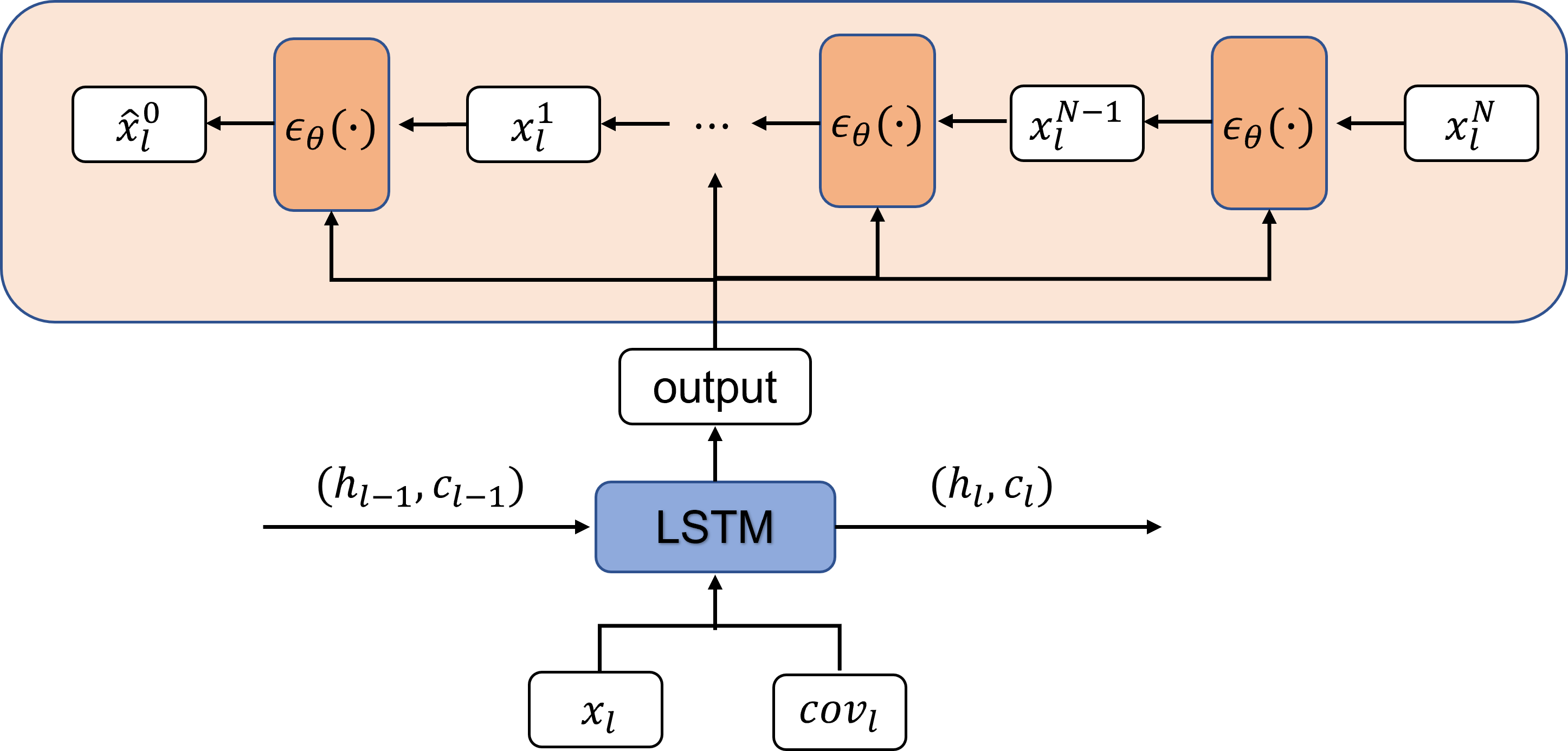}
\caption{The process of reconstructing $x_l$, i.e., the inference process of \textit{ETDddpm$_{R}$}.}
\label{fig:diffusion-r}
\end{figure}

In the following subsections, we first describe the reconstruction model \eqref{objective reconstruction} and forecasting model \eqref{objective prediction} based on DDPM, where we apply LSTM$\footnote{LSTM severs as a feature extractor in \textit{ETDddpm} considering its simple architecture and computation efficiency. LSTM can be changed to other feature extractors such as GRU \cite{dey2017gate} and transformer \cite{vaswani2017attention}.}$ as the feature extractor. Then, we construct the complete \textit{ETDddpm}.

\subsection{DDPM-based Model for Reconstruction}\label{sec:LSTM Diffusion Model for Reconstruction}

Employing DDPM, we aim to reconstruct every smart grid variable $\bm x_l \in \{\bm x_1,\bm x_2,\cdots,\bm x_L\}$ into $\bm{\hat{x}}_l^0$ starting from a Gaussian random variable input $\bm x_l^N\sim \mathcal{N}(\mathrm{\mathbf{0}},\mathrm{\mathbf{I}})$. However, for different time steps $l$, $\bm x_l$ should follow different distributions. As a result, the DDPM needs guidance to generate a suitable distribution. This guidance is the so-called conditioning input \cite{ref4}. LSTM \cite{ref5} is a popular model to extract features from time series data, which can capture not only the relationship among multiple attributes but also the dependence between the observations of different time steps. So, we use the output of an LSTM as the conditioning input. This makes the $\bm \epsilon_{\theta}(\bm x_l^n,n)$ in \eqref{ddpm objective simple 2} to be expressed as $\bm \epsilon_{\theta}(\bm x_l^n,LSTM(\bm x_{l}),n)$.


\textbf{Training process}: In the training process, firstly, we use LSTM to compute the conditioning inputs of the DDPM,
\begin{equation}
    \label{eq:conditioning input}
    \bm {con}_{1:L}, (\bm h_L,\bm c_L) = \mathrm{LSTM_{R}}(\bm x_{1:L}, \bm {cov}_{1:L}).
\end{equation}
$(\bm h_L,\bm c_L)$ is the last hidden state of LSTM which will be used as the initial hidden state in the forecasting model in Section \ref{sec:LSTM Diffusion Model for Forecasting}. For $\mathrm{LSTM_{R}}$ the initial hidden state is $\mathbf{0}$.

Then, we randomly select a diffusion step $n\in \{1,\cdots,N\}$ and generate diffusion samples according to \eqref{eq:Reparameterizing},
\begin{equation}
    \label{eq:diffusion samples}
    \bm x^n_{1:L} = \sqrt{\overline{\alpha}_n}\bm x_{1:L}+\sqrt{1-\overline{\alpha}_n}\bm\epsilon_{1:L},
\end{equation}
where $\bm\epsilon_{l}\sim \mathcal{N}(\mathrm{\mathbf{0}},\mathrm{\mathbf{I}})$ denotes the Gaussian noise added to $\bm x_l$, and we store $\bm\epsilon_{1:L}$ as labels. 

Subsequently, we estimate $\bm\epsilon_{l}$ using function $\bm \epsilon_{\theta}(\cdot)$,
\begin{equation}
    \label{eq:estimate epsilon}
    \hat{\bm \epsilon}_l = \bm \epsilon_{\theta}(\bm x_l^n,\bm {con}_l,n),\quad l\in \{1,2,\cdots,L\}.
\end{equation}
In this paper, we exploit \textit{DiffWave} \cite{ref6} as $\bm \epsilon_{\theta}(\cdot)$ with limited modification to adjust the model to our data since it has shown great performance when being utilized for DDPM to generate time series data \cite{ref6, ref7}. 

Lastly, we calculate the MSE between $\bm\epsilon_{1:L}$ and $\bm{\hat\epsilon}_{1:L}$ as the optimization objective $\mathcal{L}_R$ of Adam optimizer \cite{ref8},
\begin{equation}
    \label{eq: optimization objective r}
    \mathcal{L}_R = \mathrm{MSE}(\bm\epsilon_{1:L}-\bm{\hat\epsilon}_{1:L}).
\end{equation}
The training process ends when $\mathcal{L}_R$ converges.

\textbf{Inference process}:
In the inference process, we aim to reconstruct $\{\bm x_1^0,\bm x_2^0,\cdots,\bm x_L^0\}$ from $\{\bm x_1^N,\bm x_2^N,\cdots,\bm x_L^N\}$ in which $\bm x_l^N\sim \mathcal{N}(\mathrm{\mathbf{0}},\mathrm{\mathbf{I}})$.
Firstly, we initialize $\bm x_{1:L}^N$ and compute conditioning inputs $\bm{con}_{1:L}$ according to \eqref{eq:conditioning input}. For a given denoising step $n$, we use \eqref{eq:estimate epsilon} to estimate the Gaussian noise $\bm \epsilon_{l}^{n-1}$ that is added to $\bm x_l^{n-1}$ at the diffusion step $n-1$. With the estimated Gaussian noise $\bm {\hat\epsilon}_{l}^{n-1}$, we can estimate ${\bm x}_l^{n-1}$ according to \eqref{eq:denoising},
\begin{equation}
    \label{eq:denoising r}
    {\bm x}_l^{n-1} = \frac{1}{\sqrt{\alpha}_n}\left(\bm x_l^n-\frac{\beta_n}{\sqrt{1-\overline{\alpha}_n}}\hat{\bm\epsilon}_l^{n-1} \right) + \sigma_{\theta}\bm z.
\end{equation}
Starting from $\bm x^N_l$, we use \eqref{eq:denoising r} recurrently for $N$ steps until we obtain $\bm {\hat{x}}_l^0$, which is the desired reconstruction result. The whole inference process is shown in Fig. \ref{fig:diffusion-r}.

\subsection{DDPM-based Model for Forecasting}\label{sec:LSTM Diffusion Model for Forecasting}

For forecasting, we aim to employ DDPM to forecast $\{\bm x_{L+1},\bm x_{L+2},\cdots,\bm x_{L+T}\}$ as $\{\bm{\hat x}_{L+1}^0,\bm{\hat x}_{L+2}^0,\cdots,\bm{\hat x}_{L+T}^0\}$ with a Gaussian noise input $\bm x_{L+1:L+T}^N$.

\textbf{Training process}:
Similar to Section \ref{sec:LSTM Diffusion Model for Reconstruction}, in the training process, we first use LSTM to compute the conditioning inputs. Different from \cite{ref7}, in which the predicted $\bm{\hat x}_{L+t}$ is used as input of LSTM to compute the conditioning input $\bm{con}_{L+t+1}$ to forecast $\bm{\hat x}_{L+t+1}$, we only use $\bm {cov}_{L+t+1}$ as input of LSTM to compute $\bm{con}_{L+t+1}$ by,
\begin{equation}
    \label{eq:conditioning input f}
    \bm {con}_{L+1:L+T} = \mathrm{LSTM_F}(\bm h_L, \bm c_L,\bm {cov}_{L+1:L+T}).
\end{equation}
The initial hidden state of $\mathrm{LSTM_F}$ is $(\bm h_L,\bm c_L)$ obtained from \eqref{eq:conditioning input} which is expected to contain all the information of $\bm x_{1:L}$ and $\bm {cov}_{1:L}$. $(\bm h_L,\bm c_L)$ is the only connection between \textit{ETDddpm$_R$} and \textit{ETDddpm$_F$} as seen in Fig. \ref{fig:diffusion comb}.
This modification of the LSTM input can help to mitigate the accumulated error issue of LSTM since in the inference process, the predicted $\bm{\hat x}_{L+t}$ is usually different from $\bm{x}_{L+t}$ that is used during the training process. More importantly, with this modification, we do not need to wait for $\bm{\hat x}_{L+t}$ to compute $\bm{con}_{L+t+1}$. As a result, compared with \cite{ref7}, if we have enough computation resources, the inference speed can increase $T$ times.

Then, we randomly select a diffusion step $n$ and generate diffusion samples according to \eqref{eq:Reparameterizing},
\begin{equation}
    \label{eq:diffusion samples f}
    \bm x^n_{L+1:L+T} = \sqrt{\overline{\alpha}_n}\bm x^0_{L+1:L+T}+\sqrt{1-\overline{\alpha}_n}\bm\epsilon_{L+1:L+T},
\end{equation}
where $\bm\epsilon_{L+t}\sim \mathcal{N}(\mathrm{\mathbf{0}},\mathrm{\mathbf{I}})$, and we store $\bm\epsilon_{L+1:L+T}$ as labels. 
Subsequently, we estimate $\bm\epsilon_{L+t}$ using function $\bm \epsilon_{\theta}(\cdot)$,
\begin{equation}
    \label{eq:estimate epsilon f}
    \hat{\bm \epsilon}_{L+t} = \bm \epsilon_{\theta}(\bm x_{L+t}^n,\bm{con}_{L+t},n),\quad t\in \{1,2,\cdots,T\}.
\end{equation}

Finally we calculate the MSE between $\bm\epsilon_{L+1:L+T}$ and $\bm{\hat\epsilon}_{L+1:L+T}$ as the optimization objective $\mathcal{L}_{F}$ of Adam optimizer,
\begin{equation}
    \label{eq:optimization objective f}
    \mathcal{L}_{F} = \mathrm{MSE}(\bm\epsilon_{L+1:L+T}-\bm{\hat\epsilon}_{L+1:L+T}).
\end{equation}
The training process ends when $\mathcal{L}_{F}$ converges.

\textbf{Inference process}:
In the inference process, we aim to calculate $\{\bm x_{L+1}^0,\cdots,\bm x_{L+T}^0\}$ from $\{\bm x_{L+1}^N,\cdots,\bm x_{L+T}^N\}$ where $\bm x_{L+t}^N\sim \mathcal{N}(\mathrm{\mathbf{0}},\mathrm{\mathbf{I}})$.
First of all, we initialize $\bm x_{L+1:L+T}^N$ with Gaussian noise and compute conditioning inputs $\bm{con}_{L+1:L+T}$ according to \eqref{eq:conditioning input} and \eqref{eq:conditioning input f}. For a given denoising step $n$, we use \eqref{eq:estimate epsilon f} to estimate the Gaussian noise $\bm \epsilon_{L+t}^{n-1}$ that is added to $\bm x_{L+t}^{n-1}$ at the diffusion step $n-1$. With the estimated Gaussian noise $\bm {\hat\epsilon}_{L+t}^{n-1}$, we can estimate ${\bm x}_{L+t}^{n-1}$ according to \eqref{eq:denoising},
\begin{equation}
    \label{eq:denoising f}
    {\bm x}_{L+t}^{n-1} = \frac{1}{\sqrt{\alpha}_n}\left(\bm x_{L+t}^n-\frac{\beta_n}{\sqrt{1-\overline{\alpha}_n}}\hat{\bm\epsilon}_{L+t}^{n-1} \right) + \sigma_{\theta}\bm z.
\end{equation}
Starting from $\bm x^N_{L+t}$, we use \eqref{eq:denoising f} recurrently for $N$ steps to get $\bm {\hat{x}}_{L+t}^0$, which is the desired forecasting result. This process is similar to Fig. \ref{fig:diffusion-r}.

\subsection{Complete \textit{ETDddpm}} \label{sec:ETDddpm}
Note that \textit{ETDddpm$_R$} and \textit{ETDddpm$_F$} apply the same $\bm\epsilon_{\theta}(\cdot)$, which enforces \textit{ETDddpm$_R$} and \textit{ETDddpm$_F$} to generate the same output given the same conditioning input. This setting can also prompt the LSTMs of \textit{ETDddpm$_R$} and \textit{ETDddpm$_F$} to extract proper and consistent features.

\textbf{Training process}: According to \eqref{eq: optimization objective r} and \eqref{eq:optimization objective f}, the unified optimization objective of \textit{ETDddpm} is
\begin{equation}
    \label{eq:optimization objective comb}
    \mathcal{L}=\mathcal{L}_{R}+\gamma\mathcal{L}_{F},
\end{equation}
where $\gamma$ is a balancing coefficient, and we set it to 1 in this paper. The training process ends when $\mathcal{L}$ converges.

\textbf{Inference process:} The inference process of \textit{ETDddpm} is a simple combination of the models represented in Section \ref{sec:LSTM Diffusion Model for Reconstruction} and Section \ref{sec:LSTM Diffusion Model for Forecasting}. We can generate the reconstruction sequence $\bm{\hat x}_{1:L}^0$ according to \eqref{eq:denoising r} and generate the forecasting sequence $\bm{\hat x}_{L+1:L+T}^0$ according to \eqref{eq:denoising f}. 

The pipeline of the training and inference processes are summarized in Algorithm \ref{al:algorithm 1} and Algorithm \ref{al:algorithm 2}, respectively.

\textbf{Energy Theft Detection:} Now, considering an input sequence $\bm x_{1:L}$, we can compute the reconstruction result $\bm {\hat{x}}_{1:L}$ and the forecasting result $\bm {\hat{x}}_{L+1:L+T}$ by \textit{ETDddpm}. Subsequently, we quantify the deviation between these reconstructed and forecasted sequences and the ground truth and utilize these metrics to identify energy theft.

REMs utilize the reconstruction error \eqref{error_R} as the metric. If the reconstruction error exceeds a threshold $th_R$, we determine it as energy theft. The threshold is set manually to balance precision and recall. 
On the other hand, FEMs apply the following forecasting error as the metric,
\begin{equation}
\begin{aligned}
    \label{eq:error_F_adjust}
    \delta_F=&{\rm mean}(|\hat{\bm x}_{L+1:L+T}-\bm x_{L+1:L+T}+\\
    &{\rm mean}(\bm x_{L+1:L+T})-{\rm mean}(\hat{\bm x}_{L+1:L+T})|).
\end{aligned}
\end{equation}
We do not apply \eqref{eq:error_F} since mean shift always happens when forecasting and can significantly affect ETD performance in our experiments. With \eqref{eq:error_F_adjust}, we can ignore the mean shift and only focus on the shape of the forecasting and ground truth curves for the ETD problem. Similarly, if the forecasting error is larger than a threshold $th_F$, we identify the input as energy theft.

Considering both reconstruction and forecasting errors, we propose an ensemble method to enhance the performance of current ETD methods. Specifically, if either metric indicates an anomaly in the input, we classify it as potential energy theft. This complementary strategy ensures that in cases where one of the REM or FEM fails to detect an energy theft, the other can effectively contribute to its identification.

\begin{algorithm}[t]
\caption{Training Process of \textit{ETDddpm}}\label{al:algorithm 1}
\begin{algorithmic}
\DontPrintSemicolon
\small
\REQUIRE Randomly initialized $\epsilon_\theta(\cdot)$ and training data $X$
\ENSURE Trained $\epsilon_\theta(\cdot)$

\rule{8cm}{0.4pt}

\For{$epoch$ = $1:max\_epoch$}
{
    
    \For{each $x_{1:L+T}$ in $X$}
    {
        \tcp*[h]{\textit{\footnotesize Conditional inputs for reconstruction}}\;
        \textit{$\bm {con}_{1:L}, (\bm h_L,\bm c_L) = \mathrm{LSTM_{R}}(\bm x_{1:L}, \bm {cov}_{1:L})$ \\}        
        
        \tcp*[h]{\textit{\footnotesize Conditional inputs for forecasting}}\;
        \textit{$\bm {con}_{L+1:L+T} = \mathrm{LSTM_F}(\bm {cov}_{1:L+T})$ \\}

        \tcp*[h]{\footnotesize\textit{Randomly selected n and $\bm\epsilon_{i}\sim \mathcal{N}(\mathrm{\mathbf{0}},\mathrm{\mathbf{I}})$}}\;
        \textit{$\bm x^n_{1:L+T} = \sqrt{\overline{\alpha}_n}\bm x_{1:L+T}+\sqrt{1-\overline{\alpha}_n}\bm\epsilon_{1:L+T}$.\\}

        \tcp*[h]{\footnotesize\textit{Estimate $\bm\epsilon_{1:L+T}$ with $\epsilon_\theta(\cdot)$}}\;
        $\hat{\bm \epsilon}_{1:L+T} = \bm \epsilon_{\theta}(\bm x_{1:L+T}^n, \bm{con}_{1:L+T},n)$      

        \tcp*[h]{\footnotesize\textit{Loss function}}\;
        $\mathcal{L} = \mathrm{MSE}(\bm\epsilon_{1:L+T}-\bm{\hat\epsilon}_{1:L+T})$ 
        
        Minimizing $\mathcal{L}$ to optimize $\epsilon_\theta(\cdot)$
    }
}
\end{algorithmic}
\end{algorithm}

\begin{algorithm}[t]
\caption{Inference Process of \textit{ETDddpm}}\label{al:algorithm 2}
\begin{algorithmic}
\DontPrintSemicolon
\small

\REQUIRE Trained $\epsilon_\theta(\cdot)$ and an inference sample $\bm x_{1:L}$
\ENSURE Reconstruction and forecasting result $\bm x_{1:L+T}^0$

\rule{8cm}{0.4pt}

\tcp*[h]{\footnotesize\textit{Conditional inputs for reconstruction}}\;
\textit{$\bm {con}_{1:L}, (\bm h_L,\bm c_L) = \mathrm{LSTM_{R}}(\bm x_{1:L}, \bm {cov}_{1:L})$ \\}

\tcp*[h]{\footnotesize\textit{Conditional inputs for forecasting}}\;
\textit{$\bm {con}_{L+1:L+T} = \mathrm{LSTM_F}(\bm {cov}_{1:L+T})$ \\}

\For{n=N:1}
{
    
    $\hat{\bm \epsilon}^{n-1}_{1:L+T} = \bm \epsilon_{\theta}(\bm x_{1:L+T}^n,\bm{con}_{1:L+T},n)$

    \tcp*[h]{\footnotesize$\bm x_i^N\sim \mathcal{N}(\mathrm{\mathbf{0}},\mathrm{\mathbf{I}})$}\;
    ${\bm x}_{1:L+t}^{n-1} = \frac{1}{\sqrt{\alpha}_n}\left(\bm x_{1:L+t}^n-\frac{\beta_n}{\sqrt{1-\overline{\alpha}_n}}\hat{\bm\epsilon}_{1:L+t}^{n-1} \right) + \sigma_{\theta}\bm z$
}

\end{algorithmic}
\end{algorithm}

\section{EXPERIMENTS AND RESULTS} \label{sec:4}

In this section, we commence with a description of the datasets employed for evaluation. Subsequently, we assess the performance of \textit{ETDddpm} on ECF to provide insights into the applicability of forecasting error for energy theft detection. Following this, we present our proposed \textit{ETDddpm}-based methods along with baseline ETD methods. Finally, we compare the performance of these methods on the ETD problem, considering both regular and high-variance smart grid data.

\subsection{Datasets} \label{sec:datasets}

We employ two datasets to evaluate our proposed scheme. The first is \textit{Electricity}$\footnote{\url{https://archive.ics.uci.edu/ml/datasets/ElectricityLoadDiagrams20112014}}$ which is a real-world dataset that contains 370 customers’ hourly electricity consumption. In \textit{Electricity}, most users present a regular behaviour. The second one is \textit{Electricity-Theft}$\footnote{\url{https://github.com/asr-vip/Electricity-Theft}}$ \cite{ref11}, which is a synthetic 15-minute smart grid dataset generated with the “GridLab-D” simulation tool \cite{ref12}. 
In \textit{Electricity-Theft}, some users present a regular behaviour while some users present a medium or high-variance behaviour. Thus, we can use this dataset to evaluate ETD methods on both regular and high-variance scenarios.
In our experiments, the reconstruction length and forecasting length are both 24 hours, i.e., $L$ and $T$ are 24 samples for \textit{Electricity} and 96 samples for \textit{Electricity-Theft}.

\begin{figure}[htbp]
\centering
\includegraphics[width=\linewidth]{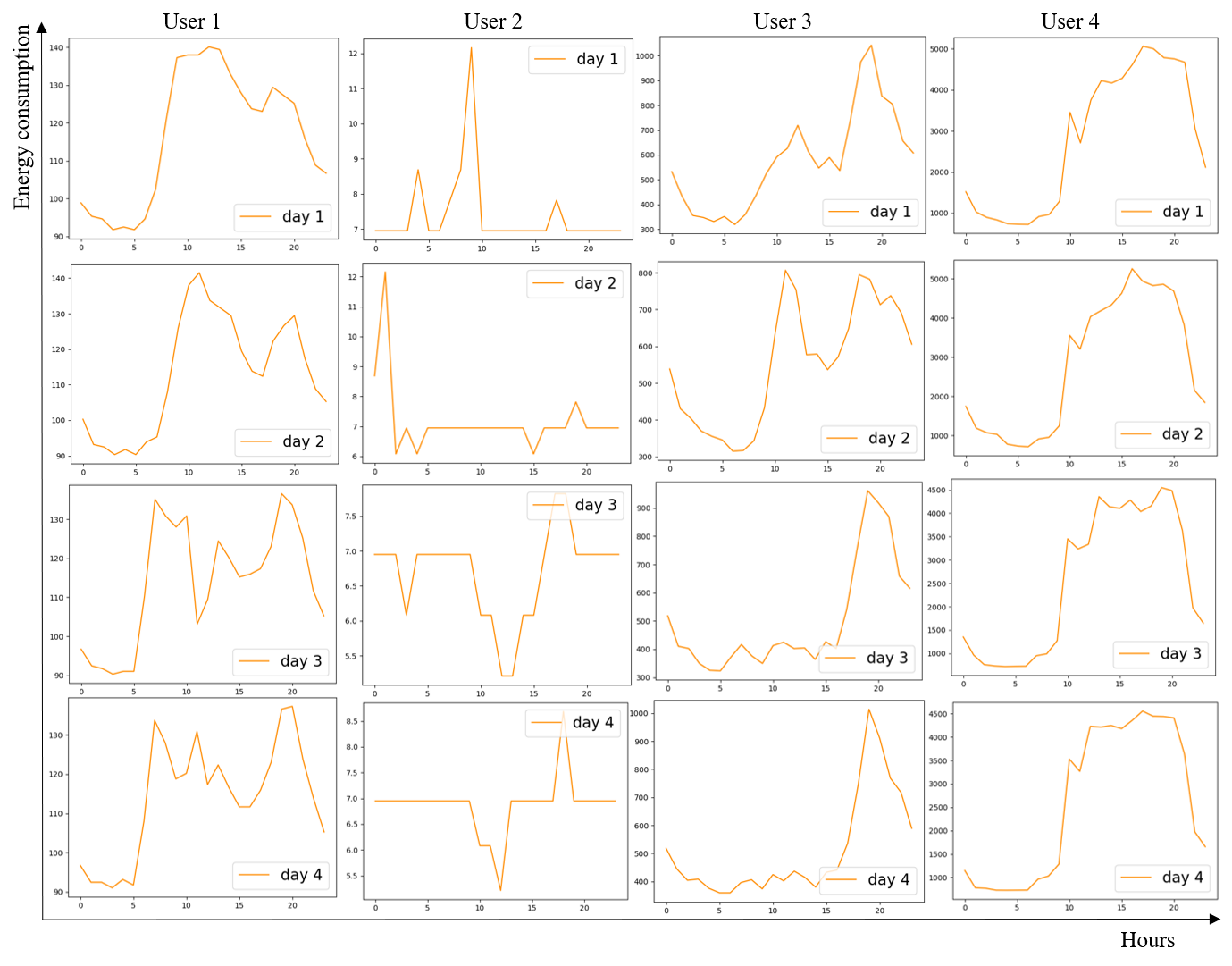}
 \caption{Illustration of 4-day energy consumption of four users in \textit{Electricity}.}
\label{fig:electricity data}
\end{figure}

\textbf{\textit{Electricity}:} 
For the user-specific scenario, we select four representative users whose electricity consumption is around ten kW$\cdot$h (user 2), a hundred kW$\cdot$h (user 1), several hundred kW$\cdot$h (user 3), and several thousand kW$\cdot$h (user 4), to construct our datasets for evaluation.  
We used the power consumption data from January 1st, 2014, to March 1st, 2014, to construct the evaluation dataset. Then, we divide the constructed dataset into three non-overlapped datasets, i.e., training (70\%), validation (10\%), and test (20\%) datasets. 
We compute the mean and the standard deviation of the training dataset and then use them to normalize all the training, validation, and test datasets. 
To evaluate the capability of the proposed and baseline ETD methods, we apply all seven types of attacks only to the test dataset since we don't need attack data to train our model. 
Figure \ref{fig:electricity data} illustrates the 4-day energy consumption of the four selected users. The figure shows that the energy consumption readings on different days show one or two similar patterns.

\begin{figure}[htbp]
\centering
\includegraphics[width=\linewidth]{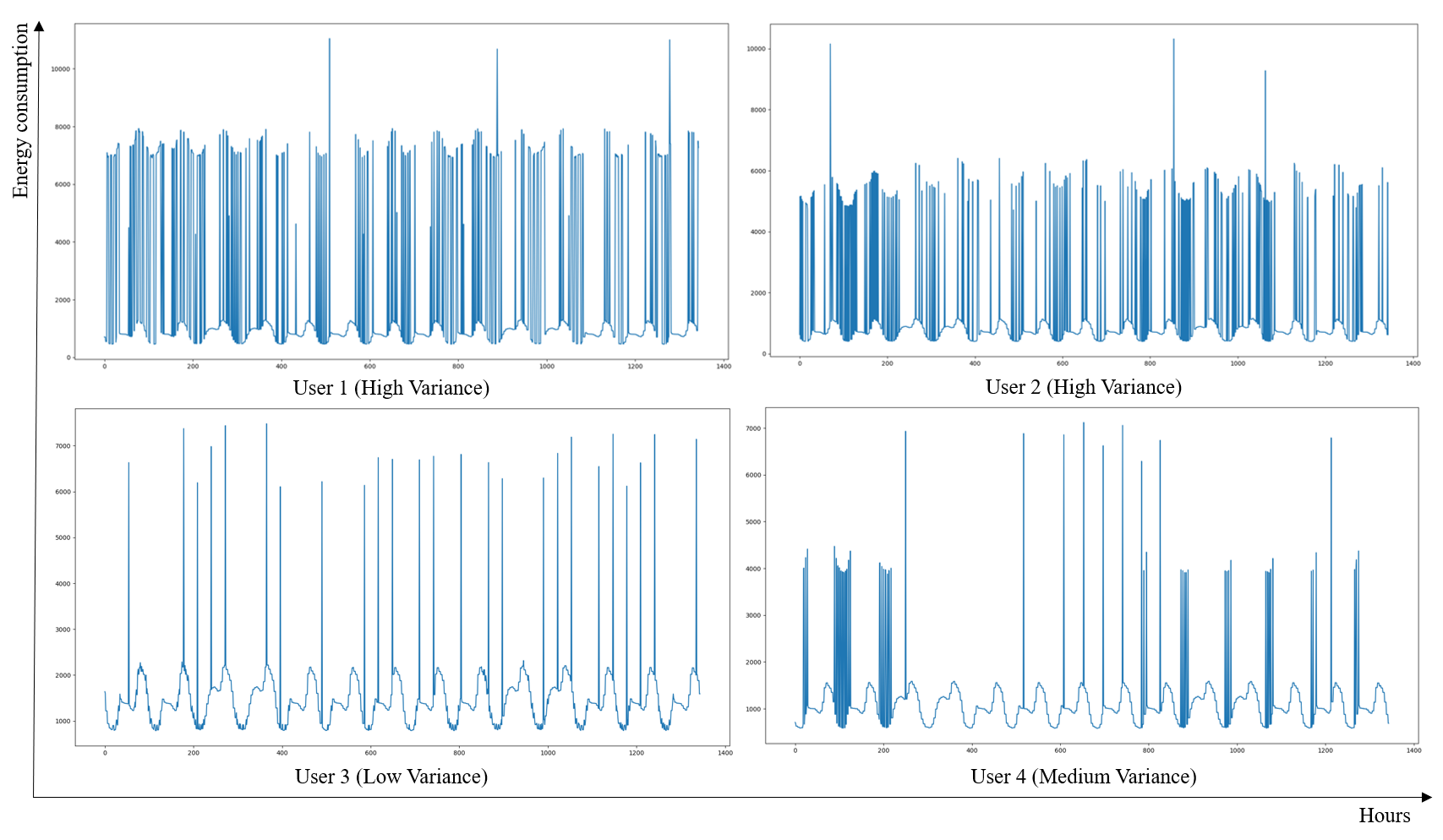}
 \caption{Illustration of 14-day energy consumption of the selected four users in \textit{Electricity-Theft}.}
\label{fig:electricity-theft data}
\end{figure}

\begin{figure}[htbp]
\centering
\includegraphics[width=0.8\linewidth]{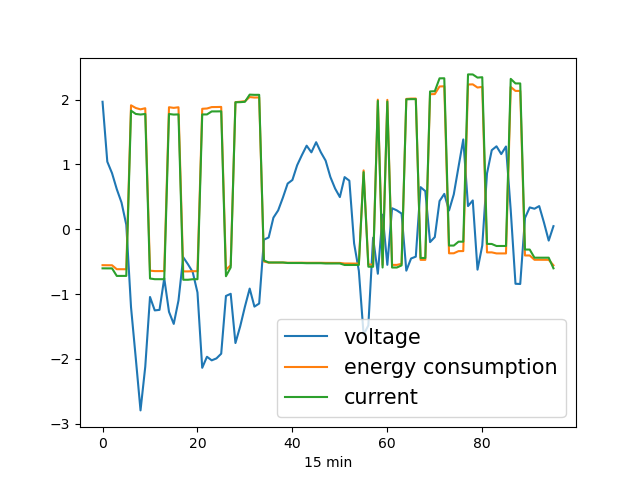}
 \caption{Illustration of the relationship among different attributes in \textit{Electricity-Theft} (all attributes undergo standardization).}
\label{fig:electricity-theft multi variables}
\end{figure}

\textbf{\textit{Electricity-Theft} \cite{ref11} :} This synthetic dataset is composed of data collected at 15-minute intervals over 31 days. 
From \textit{Electricity-Theft}, we select one user with regular energy consumption, one user with medium-variance energy consumption, and two users with high-variance energy consumption to evaluate the performance of various ETD methods in different scenarios. 
In contrast to conventional smart grid datasets like \textit{Electricity}, which solely includes energy consumption data, \textit{Electricity-Theft} encompasses both energy consumption data and additional attributes such as voltage and current. 
Given the high correlation between voltage, current, and energy consumption, our constructed dataset incorporates these attributes. 
To ensure a fair evaluation, we divide the constructed dataset into three non-overlapping subsets: training (70\%), validation (10\%), and test (20\%) datasets.
We compute the mean values and the standard deviation values of the three attributes of the training dataset, i.e., energy consumption, voltage, and current, and then use them to normalize the training, validation, and test datasets. 
Figure \ref{fig:electricity-theft data} illustrates the 14-day energy consumption data of the four selected users. We can see all users have a regular daily power consumption plus some irregular spikes simulating the scenarios where low energy-consumption devices work regularly, and high energy-consumption devices work intermittently or on demand. User 1 and User 2 in Fig. \ref{fig:electricity-theft data} present high-variance energy consumption, and User 3 and User 4 present low-variance and medium-variance energy consumption, respectively.
Figure \ref{fig:electricity-theft multi variables} illustrates normalized energy consumption, voltage, and current readings of User 1 in one day.
We can see that, after normalization, the energy consumption and current curves exhibit similar shapes.
As a result, if we only conduct attacks on `energy consumption', it can be easily detected through `current'.
To introduce a greater challenge and avoid information leakage, we apply the same attack to both `energy consumption' and `current', preserving the similarity between their curves. Additionally, all seven types of attacks are exclusively applied to the test dataset for evaluation purposes.

\subsection{Hyperparameters and Convergence Curves}

\begin{figure}[htbp]
\centering
\includegraphics[width=\linewidth]{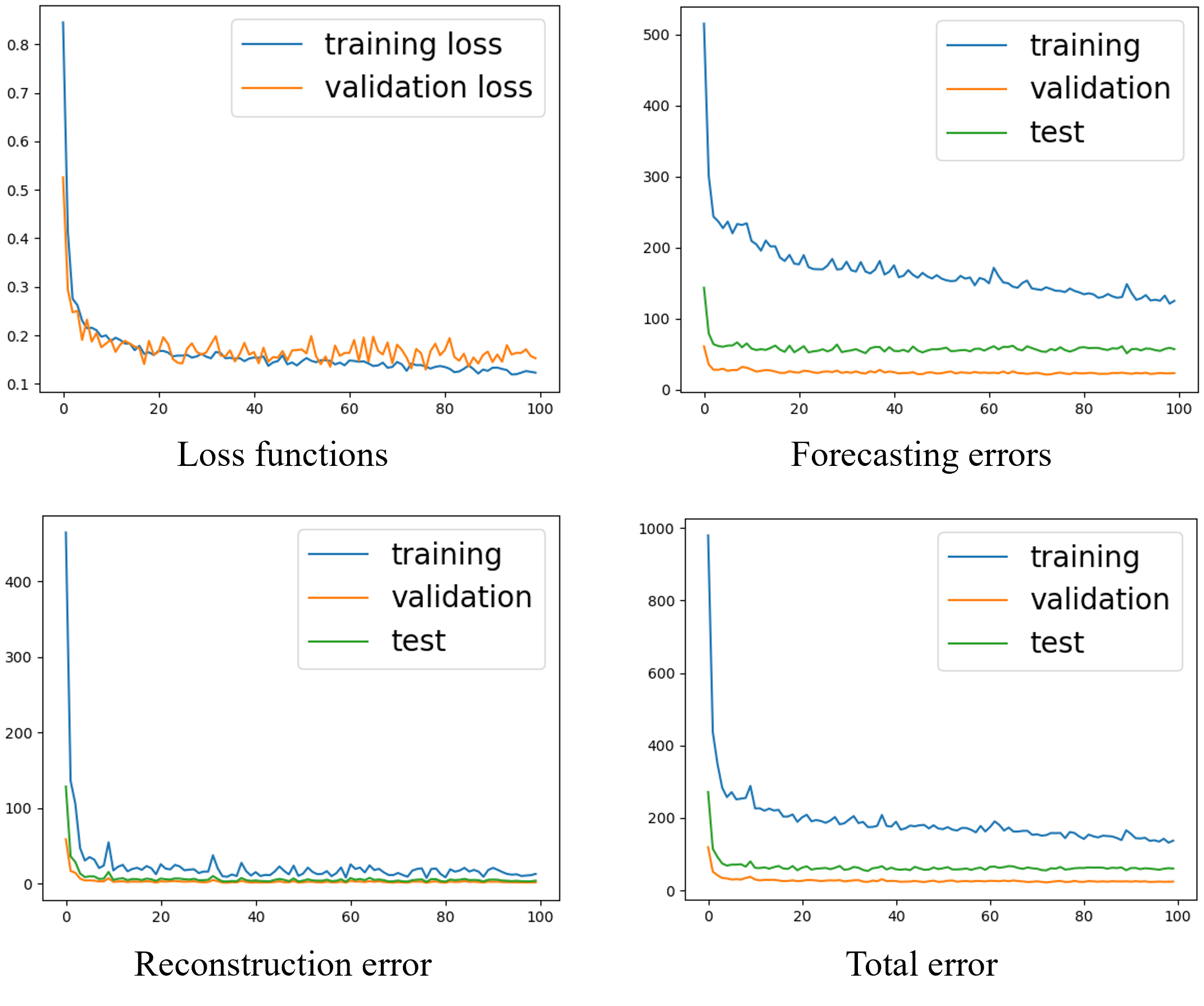}
 \caption{Convergence curves of loss function, forecasting error, reconstruction error, and total error during training iterations.}
\label{fig:convergence}
\end{figure}

We train \textit{ETDddpm} using Adam optimizer \cite{ref8} with a learning rate of 0.001. The diffusion step $N$ is set as 50. The set of variance parameters, $\beta$, is a linear variance schedule starting from $\beta_1 = 10^{-4}$ till $\beta_N = 0.05$. The training batch size is 64. In the implementation of \textit{ETDddpm}, we apply a 1-layer \textit{LSTM} as the feature extractor with hidden state $\mathbf{h}_t \in \mathbb{R}^{128}$.
The network $\epsilon_\theta(\cdot)$ consists of conditional 1-dimensional dilated ConvNets with residual connections adapted from the DiffWave \cite{ref6} model. 

Figure \ref{fig:convergence} shows the convergence curves during the training iterations of User 3 of \textit{Electricity}. We can see that all the loss function \eqref{eq:optimization objective comb}, reconstruction error \eqref{error_R}, forecasting error \eqref{error_R}, and total error \eqref{eq:unified problem} converge. The convergence shows that the reconstruction and forecasting errors of \textit{ETDddpm} are minimized by optimizing $\epsilon_\theta(\cdot)$ with the unified objective function \eqref{eq:optimization objective comb}.

\begin{figure*}[htbp]
\centering
\includegraphics[width=0.8\linewidth, scale=1.0]{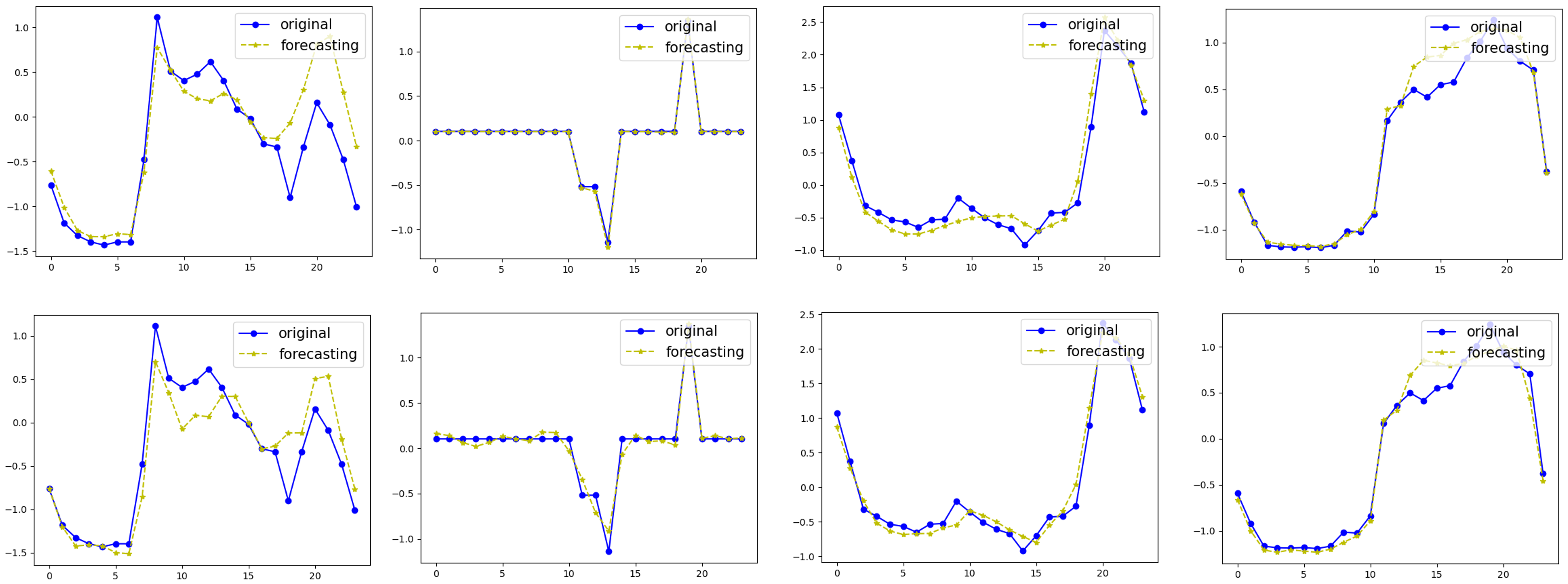}
 \caption{Forecasting examples on \textit{Electricity}. Results of the upper 4 figures are generated with \textit{ETDddpm} on user 1 to user 4 from left to right. Results of the lower 4 figures are generated with the \textit{LSTM} model on user 1 to user 4 from left to right.}
\label{fig:forecasting E}
\end{figure*}

\subsection{Experiment results on Electricity Consumption Forecasting} \label{sec:Experiment results on ECF}

In this section, we evaluate the proposed \textit{ETDddpm} on the ECF problem. 
Since ECF is not the focus of this study, we only apply LSTM as the baseline for comparison.
Table \ref{tab:forecasting electricity} shows the MAE of the two methods on \textit{Electricity} and \textit{Electricity-Theft}. 
For the \textit{Electricity} dataset, we calculate the mean absolute error on the normalized energy consumption data of each time step.
We observe that the performance of \textit{ETDddpm} is comparable to that of the \textit{LSTM} model.
We also provide some visualization results on \textit{Electricity} in Fig. \ref{fig:forecasting E}. 
From Table \ref{tab:forecasting electricity}, we can see that both methods cannot perform satisfactorily on User 1. 
In the leftmost column of Fig. \ref{fig:forecasting E}, both methods tend to forecast higher values. 
This observation is consistent with the data characteristics, i.e., the values of test data are consistently smaller than those of training and validation data. In the general time series forecasting area \cite{ref7}, instance normalization \cite{ulyanov2016instance} is usually used to avoid this problem, i.e., mean shift between training and test data. However, in the energy theft detection scenario, `fixed reduction' and `partial reduction' attacks cannot be detected if the input is preprocessed with instance normalization because the normal and artificial sequences will become identical. Fortunately, although the MAE is relatively high on User 1 of \textit{Electricity}, the shapes of the forecasting curves and the ground truths are similar. Thus, we can distinguish the normal and the attack sequences by the forecasting error computed by equation \eqref{eq:error_F_adjust} that eliminates the impact of changes in the mean value. 

\begin{figure}[htbp]
\centering
\includegraphics[width=0.9\linewidth]{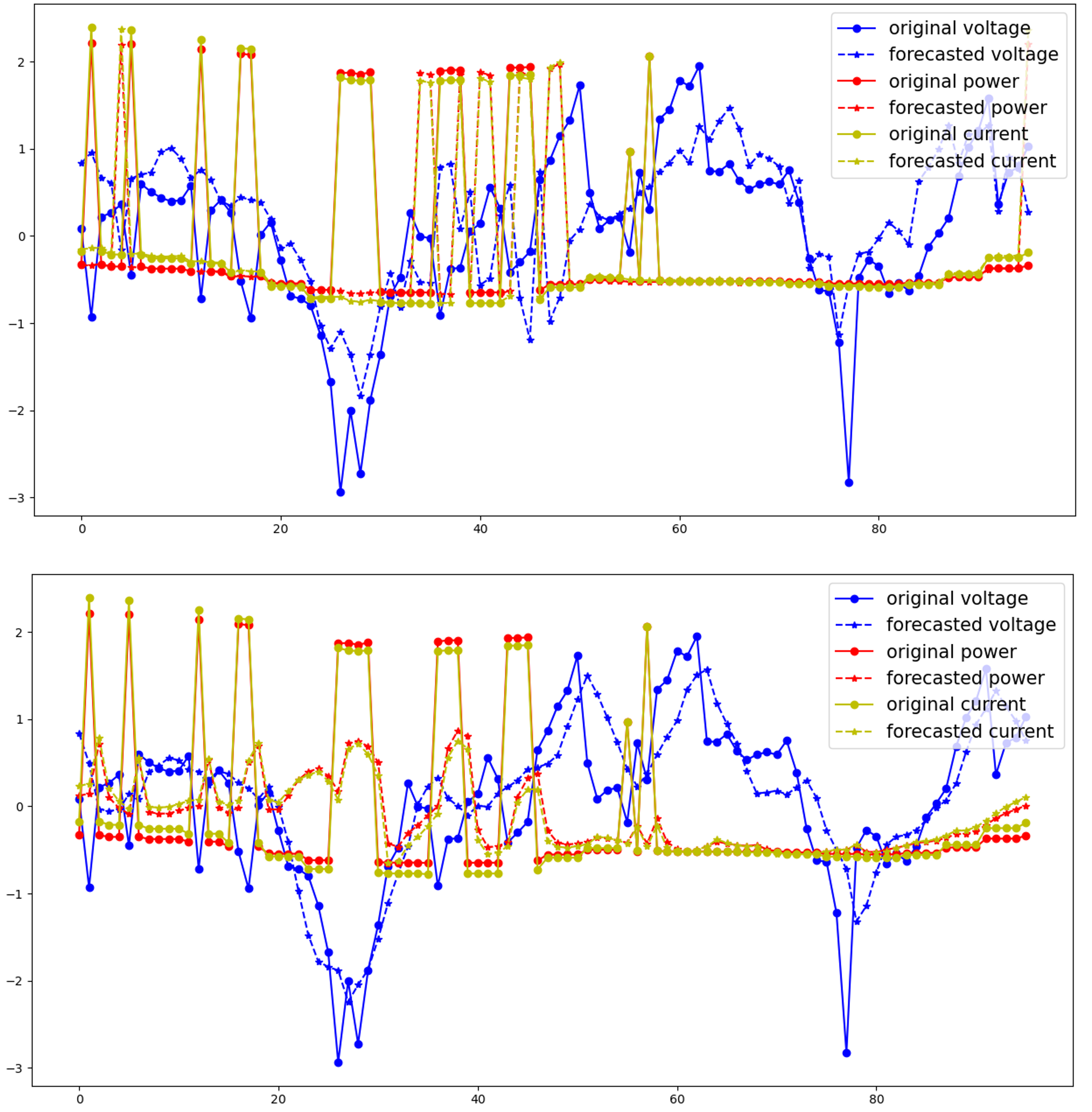}
 \caption{Forecasting examples on the datasets constructed with \textit{Electricity-Theft}. The result on the upper figure is generated with \textit{ETDddpm}. The result of the lower figure is generated with the \textit{LSTM} model.}
\label{fig:forecasting ET}
\end{figure}

On the other hand, for the \textit{Electricity-Theft} dataset, we calculate the mean absolute error on the normalized `energy consumption', `voltage', and `current' at each time step in Table \ref{tab:forecasting electricity}. 
Figure \ref{fig:forecasting ET} shows the forecasting results of ETDddpm and the LSTM on User 1 of \textit{Electricity-Theft}. We can see that \textit{ETDddpm} and the \textit{LSTM} model show different behaviours. The forecasting sequence of \textit{ETDddpm} shows high variance on `energy consumption' and `current' to approximate the ground-truth behaviour while that of \textit{LSTM} tries to predict the expectations of the `energy consumption' and `current'. However, it is evident that both LSTM and \textit{ETDddpm} exhibit incapacity in forecasting the energy consumption of the user with high variance from Table \ref{tab:forecasting electricity} and Fig. \ref{fig:forecasting ET}.
As a result, Assumption 2 mentioned in Section \ref{problem statement} is compromised due to the high forecasting error, which can lead to bad performance on the ETD problem. 

Fortunately, for the ETD problem, we know the true input and future data, i.e., $\bm x_{1:L+T}$. Thus, we do not need to forecast future energy consumption from Gaussian noise, i.e., $\bm x^N_{1:L+T}$, using \textit{ETDddpm}. Instead, we can conduct the diffusion procedure for $N_1 < N$ times and input $\bm x^{N_1}_{1:L+T}$ into \textit{ETDddpm} for reconstruction and forecasting, and this method is named \textit{ETDddpm}$^+$. In our implementation, $N$ is 50 and $N_1$ is 20. 
After 20 diffusion steps, $\bm x^{20}_{1:L+T}$ preserves partial information of the original sequence, $\bm x_{1:L+T}$, in contrast to Gaussian noise $\bm x^{50}_{1:L+T}$ that eliminates all information. Thus, $\bm x^{20}_{1:L+T}$ provides a good starting point for the denoising process of DDPM, which can mitigate the uncertainty caused by the high-variance input data.
As shown in Table \ref{tab:forecasting electricity}, \textit{ETDddpm}$^+$ greatly improves the forecasting accuracy. As a result, Assumption 2 mentioned in Section \ref{problem statement} is held with the help of \textit{ETDddpm}$^+$ and we can expect a good ETD performance employing \textit{ETDddpm}$^+$.

\begin{figure}[htbp]
\centering
\includegraphics[width=0.8\linewidth, scale=1.]{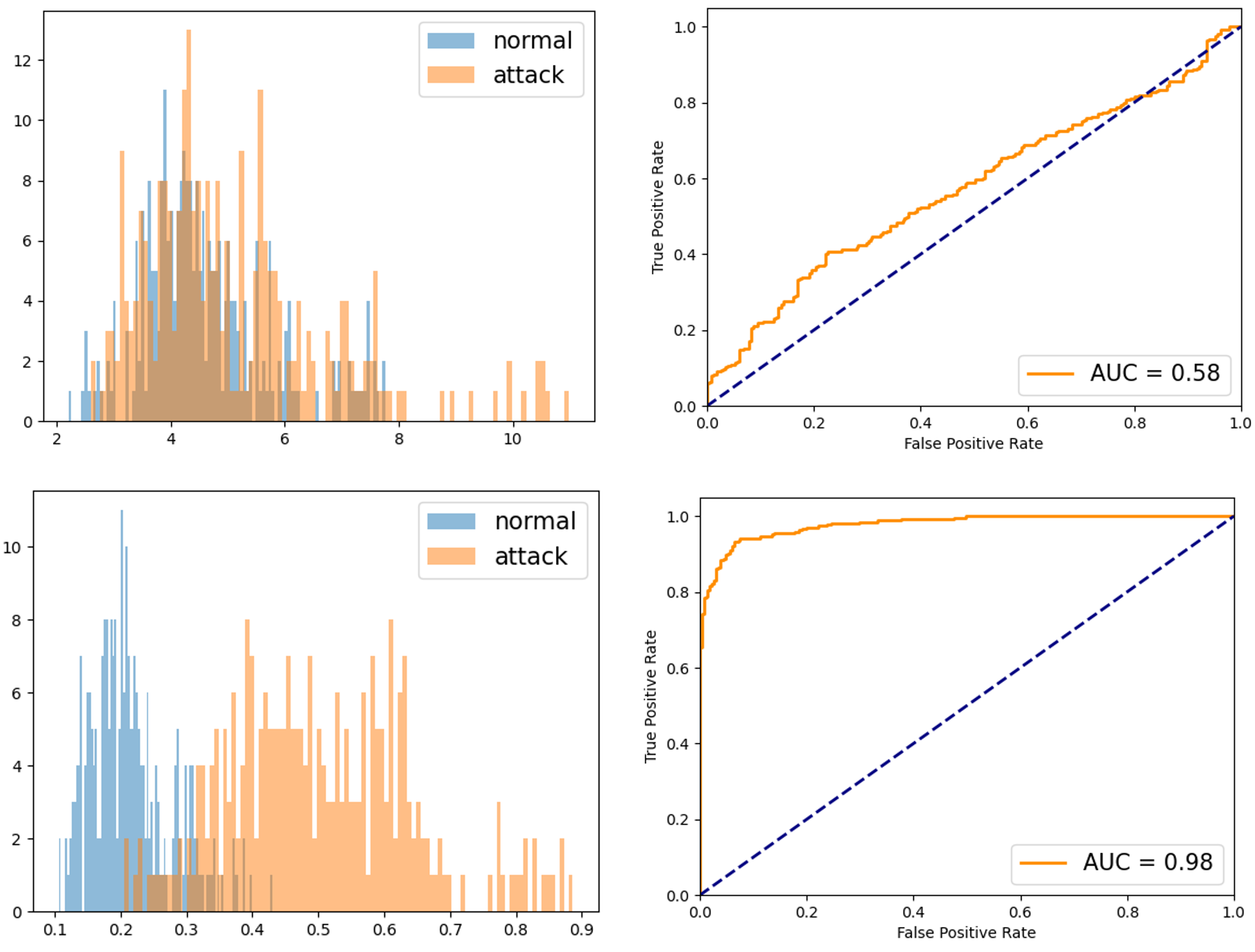}
\caption{This figure illustrates the ETD performance on User 3 under a fixed reduction attack. The upper left figure employs forecasting error as the anomaly score, where blue bins present anomaly scores of normal sequences and orange bins present anomaly scores of attack sequences. The upper right figure shows the ROC curve and AUC based on the forecasting anomaly score. The two bottom figures are similar to the two upper figures but employ reconstruction error as the anomaly score.}
\label{fig:example-etd-electricity}
\end{figure} 

\begin{table}[htbp]
\renewcommand\arraystretch{1.0}
    \centering
    \caption{MAE of Different ECF Methods on \textit{Electricity} and \textit{Electricity-Theft}}
    \begin{tabular}{c|c|c|c|c|c}
    \hline
        \textbf{\textit{Electricity}} & \textbf{User 1} & \textbf{User 2}& \textbf{User 3}&\textbf{User 4}&\textbf{Ave}\\
         \hline
         {\textbf{LSTM}\cite{ref38}} &{0.292}&0.077&{0.155}&0.083 &\textbf{\textcolor{red}{0.219}}\\
         \hline
         {\textbf{ETDddpm}} &0.633&{0.016}&0.203 &{0.082} &0.241 \\
         \hline

         \textbf{\textit{Electricity-Theft}} & \textbf{User 1} & \textbf{User 2}& \textbf{User 3}&\textbf{User 4}&\textbf{Ave}\\
         \hline
         {\textbf{LSTM}\cite{ref38}} &0.660 &0.676 &0.402 &0.396 &0.534\\
         \hline
         {\textbf{ETDddpm}} &0.642 &0.634 &0.348 &0.323 &0.487 \\
         \hline
         {\textbf{ETDddpm$^+$}} &0.181 &0.170 &0.186 &0.146 &\textbf{\textcolor{red}{0.171}} \\
         \hline
    \end{tabular}
    \label{tab:forecasting electricity}
\end{table}

\subsection{Experiment Results on Electricity Theft Detection} \label{sec:Experiment results on ETD}

In this section, we begin by introducing the evaluation metrics. Second, we introduce the proposed and baseline ETD methods. Then, we present the performance results of various ETD methods on \textit{Electricity} and \textit{Electricity-Theft}. Finally, we demonstrate the enhanced performance achieved through the ensemble method, highlighting its superiority over individual REM and FEM approaches. In this section, we first show experimental results on the user-specific scenario and experimental results on the multiple-user scenario are shown in Section \ref{sec:scalability}.

\subsubsection{Evaluation Metrics}

\begin{table*}[ht]
\renewcommand\arraystretch{1.0}
    \caption{AUC Scores of Different ETD Methods on \textit{Electricity} and \textit{Electricity-Theft}}
    \label{tab:AUC scores user-specific}
    \centering
    \begin{tabular}{c|c|c|c|c|c|c|c|c|c|c|c|c|c|c|c|c}
    \hline
    {} & \multicolumn{8}{c|}{\textbf{\textit{Electricity}}} & \multicolumn{8}{c}{\textbf{ \textit{Electricity-Theft}}}\\
    \hline
    
        {} &\makebox[0.024\textwidth]{\textbf{FR}} & \makebox[0.024\textwidth]{\textbf{PR}} & \makebox[0.024\textwidth]{\textbf{RPR}} & \makebox[0.024\textwidth]{\textbf{SBP}} &\makebox[0.024\textwidth]{\textbf{AC}} & \makebox[0.024\textwidth]{\textbf{RAC}} & \makebox[0.024\textwidth]{\textbf{REV}} & \makebox[0.024\textwidth]{\textbf{Ave}}&\makebox[0.024\textwidth]{\textbf{FR}} & \makebox[0.024\textwidth]{\textbf{PR}} & \makebox[0.024\textwidth]{\textbf{RPR}} & \makebox[0.024\textwidth]{\textbf{SBP}} &\makebox[0.024\textwidth]{\textbf{AC}} & \makebox[0.024\textwidth]{\textbf{RAC}} & \makebox[0.024\textwidth]{\textbf{REV}} & \makebox[0.024\textwidth]{\textbf{Ave}}\\
         \hline
         {} & \multicolumn{8}{c|}{\textbf{User 1 (Regular Pattern)}} & \multicolumn{8}{c}{\textbf{User 1 (High-Variance)}} \\
    \hline
         {\textbf{L-R}\cite{ref37}} &0.93 & 0.93 & 0.94 & 1.00 &0.09 &0.40&0.51&0.69    &0.61 &0.27 &0.26 &0.23 &0.01 &0.01 &0.58 &0.28\\
         \hline
         {\textbf{L-F}\cite{ref38}} &0.89 &0.86 &0.86 &1.00 &0.13 &0.68 &0.52 & 0.71          &0.71 &0.44 &0.43 &0.94 &0.06 &0.03 & 0.58 &0.46\\
         \hline
         {\textbf{FC-R}\cite{ref9}} &1.00 &1.00 &1.00 &1.00 &0.00 &1.00 &0.52 &0.79      &0.85 &0.40 &0.38 &1.00 &0.00 &0.00 &0.60 &0.46\\
         \hline
         {\textbf{VAE-R}\cite{ref10}} &0.99 &0.98& 0.98& 1.00&0.36&1.00&0.56&\textbf{\textcolor{blue}{0.84}}    &0.83 &0.41 & 0.39 & 0.99 &0.00 &0.00 &0.66 &0.47\\
         \hline
         {\textbf{ED-R (ours)}} &0.99&0.98&0.98&1.00&0.13&0.45&0.51&0.72 
         &1.00&1.00&1.00&1.00&1.00&1.00&1.00&\textbf{\textcolor{red}{1.00}}\\
         \hline
         {\textbf{ED-F (ours)}} &0.82&0.80&0.88&1.00&0.99&1.00&1.00&\textbf{\textcolor{red}{0.93}}
         &0.87&0.86&0.86&0.98&1.00&1.00&0.92&\textbf{\textcolor{blue}{0.93}}\\

    \hline
    {} & \multicolumn{8}{c|}{\textbf{User 2 (Regular Pattern)}} & \multicolumn{8}{c}{\textbf{User 2 (High-Variance)}} \\
    \hline
    {\textbf{L-R}\cite{ref37}} &1.00 &1.00 &1.00 &0.05 &0.01 &1.00 &0.93 &0.71           &0.72 &0.40 &0.39 &0.83 &0.00 &0.00 &0.62 &0.42\\
     \hline
     {\textbf{L-F}\cite{ref38}} &1.00&1.00&1.00&1.00&1.00&1.00&1.00&\textbf{\textcolor{red}{1.00}}
     &0.77 &0.49 &0.48 &0.95 &0.00 &0.01 &0.50 &0.46\\
     \hline
     {\textbf{FC-R}\cite{ref9}} &1.00 &1.00 &1.00 &1.00 &0.00 &1.00 &0.63 &0.80     &0.83 &0.40 &0.39 &0.97 &0.00 &0.00 &0.53 &0.45\\
     \hline
     {\textbf{VAE-R}\cite{ref10}} &1.00 &1.00& 1.00& 0.00&0.00&1.00&0.49&0.64        &0.85 &0.40 & 0.39 & 0.99 &0.00 &0.01 &0.61 &0.47\\
         \hline
     {\textbf{ED-R (ours)}} &1.00&1.00&1.00&1.00&1.00&1.00&0.86 &0.98
     &1.00 &0.85&0.86&1.00&1.00&1.00&0.98& \textbf{\textcolor{blue}{0.96}}\\
     \hline
    {\textbf{ED-F (ours)}} &0.91&0.99&1.00&1.00&1.00&1.00&1.00 &\textbf{\textcolor{blue}{0.99}}
    &0.94 &0.92&0.92&0.98&1.00&1.00&1.00& \textbf{\textcolor{red}{0.97}}\\

    \hline
    {} & \multicolumn{8}{c|}{\textbf{User 3 (Regular Pattern)}} & \multicolumn{8}{c}{\textbf{User 3 (Regular Pattern)}} \\
    \hline
    {\textbf{L-R}\cite{ref37}} &0.96 &0.89 &0.94 &1.00 &0.98 &0.98 &0.96 &\textbf{\textcolor{blue}{0.96}}     &1.00 &0.99 &1.00 &1.00 & 0.99 &0.97 &1.00 &0.99\\
     \hline
     {\textbf{L-F}\cite{ref38}} &0.96&0.94&0.94 &1.00 &0.99 &0.97 &0.83 &0.95         &1.00 &1.00 &1.00 &1.00 &0.99 &1.00 &1.00 &\textbf{\textcolor{red}{1.00}}\\
     \hline
     {\textbf{FC-R}\cite{ref9}} &0.96 &0.93 &0.92 &1.00 &0.00 &0.33 &0.55 &0.67          &1.00 &0.59 &0.62 &1.00 &0.01 &0.12 &0.70 &0.58\\
     \hline
     {\textbf{VAE-R}\cite{ref10}} &0.94 &0.81& 0.89 & 1.00 &0.99 &0.98 &0.95 &0.94       &1.00 &0.98 & 0.99 & 1.00 &1.00 &1.00 &1.00 &\textbf{\textcolor{red}{1.00}}\\
     \hline
     {\textbf{ED-R (ours)}} &0.98&0.95&0.96&1.00&0.97&0.93&1.00&\textbf{\textcolor{red}{0.97}}
      &1.00 &0.99 &1.00 &1.00 & 0.21 &0.17 &1.00 & {0.77}\\
     \hline
     {\textbf{ED-F (ours)}} &0.58&0.62&0.83&1.00&1.00&1.00&1.00&0.86
      &1.00 &1.00 &1.00 &1.00 & 1.00 &1.00 &1.00 & \textbf{\textcolor{red}{1.00}} \\

    \hline
    {} & \multicolumn{8}{c|}{\textbf{User 4 (Regular Pattern)}} & \multicolumn{8}{c}{\textbf{User 4 (Medium-Variance)}} \\
    \hline
     {\textbf{L-R}\cite{ref37}} &1.00 &1.00 &1.00 &1.00&1.00&1.00&1.00& \textbf{\textcolor{red}{1.00}}          &1.00 &0.75 &0.75 &1.00 &0.68 &0.85 &0.94 &0.85\\
     \hline
     {\textbf{L-F}\cite{ref38}} &0.94&1.00&1.00&0.99&1.00&1.00&1.00&0.99
     &1.00 &0.94 &0.93 &1.00 &0.52 &0.79 &0.93 &0.87\\
     \hline
     {\textbf{FC-R}\cite{ref9}} &1.00 &1.00 &1.00 &1.00 &0.00 &1.00 &0.86 &0.84     &1.00&0.57&0.58&1.00 &0.03 &0.15 &0.65 &0.57\\
     \hline
     {\textbf{VAE-R}\cite{ref10}} &1.00 &1.00& 1.00 & 1.00 &1.00 &1.00 &1.00 &\textbf{\textcolor{red}{1.00}}     &1.00 &0.96 & 0.96 & 1.00 &0.74 &0.91 &0.94 &0.93\\
     \hline
     {\textbf{ED-R (ours)}} &1.00&0.99&0.99&1.00&1.00&0.99&1.00&\textcolor{red}{\textbf{1.00}}     &1.00 &1.00 &1.00 &1.00 & 1.00 &1.00 &0.99 & \textbf{\textcolor{red}{1.00}}\\
     \hline
     {\textbf{ED-F (ours)}} &0.72&1.00&1.00&1.00&1.00&1.00&1.00&0.96   &1.00 &1.00 &1.00 &1.00 & 1.00 &1.00 &1.00 & \textbf{\textcolor{red}{1.00}}\\
         
         \hline
    \end{tabular}
    \begin{tablenotes}
        \footnotesize
        \item\textbf{FR}: Fixed reduction; \textbf{PR}: Partial reduction; \textbf{RPR}: Random partial reduction; \textbf{SBP}: Selective by-pass;
        \textbf{AC}: Average consumption; \textbf{RAC}: Random average consumption; \textbf{REV}: Reverse; \textbf{Ave}: average.
        \textbf{ED-R}: ETDddpm-R; \textbf{ED-F}: ETDddpm-F; \textbf{L-R}: LSTM-R; \textbf{L-F}: LSTM-F.
    \end{tablenotes}
\end{table*}

When evaluating the performance of ETD methods on a specific energy theft attack, we generate an attack sequence for each normal sequence of the test dataset. Since the number of normal and attack samples is the same, there is no need to use the precision-recall curve, which is developed for highly imbalanced test datasets. Instead, we utilize the receiver operating characteristic (ROC) curve for evaluation, which is a graph showing the performance of a classification model at all classification thresholds.
The y-axis of an ROC curve denotes the true positive rate (TPR) and the x-axis denotes the false positive rate (FPR). TPR is a synonym for recall and is therefore defined as, $TPR=\frac{TP}{TP+FN}$, where $TP$ denotes the number of true positive samples and $FN$ denotes the number of false negative samples. FPR is defined as $FPR=\frac{FP}{FP+TN}$, where $FP$ denotes the number of false positive samples and $TN$ denotes the number of true negative samples. 
The Area Under the ROC Curve (AUC), ranging from 0 to 1, serves as a metric to gauge the efficacy of ETD in this paper. AUC equal to 1 indicates near-perfect discrimination between positive and negative samples. AUC around 0.5 suggests that anomaly scores for positive and negative samples share a similar distribution, making them indistinguishable. AUC less than 0.5 signifies that positive sample scores are typically lower than negative ones. While in some general problems, AUC around 0 can be easily transformed to AUC around 1 by interchanging the definitions of positive and negative samples, such an interchange is not permissible for the ETD problem as per the assumptions outlined in Section \ref{problem statement}. Consequently, an AUC around or below 0.5 indicates a complete failure of the method.
As the AUC encompasses all conceivable thresholds and considers both TPR and FPR, the necessity to employ evaluation metrics like accuracy, precision, recall, and F1 score—limited to specific thresholds—is obviated.

However, for the ensemble model, the incorporation of two distinct thresholds, namely $th_R$ and $th_F$ for the REM and FEM, poses challenges in generating a conventional ROC curve for comprehensive evaluation. To assess the performance enhancement brought about by the ensemble method in addressing the ETD problem, we introduce a novel evaluation metric, denoted as $\alpha$\text{-}\textit{TPR}, which characterizes the TPR while constraining the FPR to a maximum threshold, $\alpha$. With a consistent FPR, an elevated TPR (recall) correlates with increased precision, accuracy, and F1 score. Consequently, $\alpha$\text{-}\textit{TPR} singularly serves as a sufficient metric for evaluating the performance of diverse ETD methods. In summary, individual REMs or FEMs are evaluated using the AUC, while the ensemble method's performance is assessed through $\alpha$\text{-}\textit{TPR}.

\begin{table*}[htbp]
\renewcommand\arraystretch{1.0}
    \centering
    \caption{True Positive Rate of Different ETD Methods on \textit{Electricity} under 5\% and 10\% False Positive Rate}
    \label{tab:AUC scores of single models and the ensemble model}
    \begin{tabular}{c|c|c|c|c|c|c|c|c|c|c|c|c|c|c|c|c}
    \hline

    
        {} &\makebox[0.026\textwidth]{\textbf{FR}} & \makebox[0.026\textwidth]{\textbf{PR}} & \makebox[0.026\textwidth]{\textbf{RPR}} & \makebox[0.026\textwidth]{\textbf{SBP}} &\makebox[0.026\textwidth]{\textbf{AC}} & \makebox[0.026\textwidth]{\textbf{RAC}} & \makebox[0.026\textwidth]{\textbf{REV}} & \makebox[0.026\textwidth]{\textbf{Ave}} &\makebox[0.026\textwidth]{\textbf{FR}} & \makebox[0.026\textwidth]{\textbf{PR}} & \makebox[0.026\textwidth]{\textbf{RPR}} & \makebox[0.026\textwidth]{\textbf{SBP}} &\makebox[0.026\textwidth]{\textbf{AC}} & \makebox[0.026\textwidth]{\textbf{RAC}} & \makebox[0.026\textwidth]{\textbf{REV}} & \makebox[0.026\textwidth]{\textbf{Ave}}\\
         \hline
         {} & \multicolumn{8}{c|}{\textbf{User 1 (Regular Pattern) under 5\% FPR}} & \multicolumn{8}{c}{\textbf{User 1 (Regular Pattern) under 10\% FPR}} \\
    \hline
         {\textbf{L-R}\cite{ref37}} &0.80 & 0.84 & 0.82 & 1.00 &0 &0.05 &0.08 &0.51    &0.85 & 0.88 & 0.88 & 1.00 &0 &0.09 &0.13 &0.55\\
         \hline
         {\textbf{L-F}\cite{ref38}} &0.09 &0.02 &0.03 &1.00 &0 &0 &0.05 & 0.17
         &0.14 &0.08 &0.08 &1.00 &0 &0 &0.09 & 0.20\\
         \hline
         {\textbf{FC-R}\cite{ref9}} &1.00 &1.00 &1.00 &1.00 &0.00 &1.00 &0.10 &0.73  &1.00 &1.00 &1.00 &1.00 &0.00 &1.00 &0.15 &0.74\\
         \hline
         {\textbf{VAE-R}\cite{ref10}} &0.97 &0.97& 0.96& 1.00&0 &1.00 &0.09 &0.71
         &1.00 &0.99& 0.99& 1.00&0 &1.00 &0.15 &0.73\\
         \hline
         {\textbf{ED-R (ours)}} &0.97 &0.94 &0.95 &1.00 &0.02 &0.02 &0.12 &0.57 & 0.98 & 0.94 & 0.95 & 1.00 & 0.02 & 0.02 & 0.15 & 0.58\\
         \hline
         {\textbf{ED-F (ours)}} &0.68 &0.65 &0.68 &1.00 &1.00 &1.00 &1.00 &\textbf{\textcolor{blue}{0.86}} & 0.69 & 0.68 & 0.72 & 1.00 & 1.00 & 1.00 & 1.00 & \textbf{\textcolor{blue}{0.87}}\\
         \hline
         {\textbf{ED-E (ours)}} &0.98 &0.98 &0.98 &1.00 &0.98 &0.99 &1.00 &\textbf{\textcolor{red}{0.99}} & 0.98 & 0.98 & 0.98 & 1.00 & 1.00 & 1.00 & 1.00 &\textbf{\textcolor{red}{0.99}}\\
         \hline
    
    {} & \multicolumn{8}{c|}{\textbf{User 3 (Regular Pattern) under 5\% FPR}} & \multicolumn{8}{c}{\textbf{User 3 (Regular Pattern) under 10\% FPR}} \\
         \hline
         {\textbf{L-R}\cite{ref37}} &0.74 & 0.37 & 0.45 & 1.00 &1.00 &0.97 &0.78 &0.76    &0.96 & 0.78 & 0.91 & 1.00 &1.00 &1.00 &1.00 &\textbf{\textcolor{red}{0.95}}\\
         \hline
         {\textbf{L-F}\cite{ref38}} &0.83 &0.71 &0.72 &1.00 &1.00 &0.92 &0.33 & 0.79      &0.89 &0.79 &0.80 &1.00 &1.00 &0.95 &0.37 & 0.83\\
         \hline
         {\textbf{FC-R}\cite{ref9}} &0.92 &0.89 &0.91 &1.00 &0 &0 &0.11 &0.55
         &0.93 &0.91 &0.91 &1.00 &0 &0 &0.19 &0.56\\
         \hline
         {\textbf{VAE-R}\cite{ref10}} &0.47 &0.14& 0.16& 1.00& 1.00 &1.00 &0.44 &0.60    &0.94 &0.56& 0.74& 1.00& 1.00 &1.00 &1.00 &0.89\\
         \hline
         {\textbf{ED-R (ours)}} &0.90 &0.71 &0.75 &1.00 &0.81 &0.78 &1.00 &\textbf{\textcolor{blue}{0.85}}      
         &0.94 &0.85 &0.86 &1.00 &0.92 &0.71 &1.00 &0.90\\
         \hline
         {\textbf{ED-F (ours)}} &0.11 &0.08 &0.18 &1.00 &1.00 &1.00 &1.00 &0.62
         &0.22 &0.15 &0.44 &1.00 &1.00 &1.00 &1.00 &0.68\\
         \hline
         {\textbf{ED-E (ours)}} &0.86 &0.62 &0.67 &1.00 &1.00 &1.00 &1.00 &\textbf{\textcolor{red}{0.88}}
         &0.92 &0.76 &0.83 &1.00 &1.00 &1.00 &1.00 &\textbf{\textcolor{blue}{0.93}}\\
         \hline
    \end{tabular}
    \begin{tablenotes}
        \footnotesize
        \item Abbreviations are the same as Table \ref{tab:AUC scores user-specific}; \textbf{ED-E}: ETDddpm-E.
    \end{tablenotes}
\end{table*}

\subsubsection{Proposed and Baseline ETD Methods}
Now, we introduce our proposed ETD methods, i.e., \textit{ETDddpm\text{-}R}, \textit{ETDddpm\text{-}F}, and \textit{ETDddpm\text{-}E}, and baseline methods, including \textit{LSTM\text{-}R} \cite{ref37}, \textit{LSTM\text{-}F} \cite{ref38}, \textit{FC\text{-}R} \cite{ref9}, and \textit{VAE\text{-}R} \cite{ref10}.

In \textit{ETDddpm\text{-}R}, we leverage the \textit{ETDddpm} introduced in Section \ref{sec:ETDddpm} to generate the reconstruction of an input sequence. We then calculate the reconstruction error \eqref{error_R} as the anomaly score.
In \textit{ETDddpm\text{-}F}, we utilize \textit{ETDddpm} to generate the forecasted sequence of an input sequence. We then calculate the forecasting error \eqref{eq:error_F_adjust} to derive the anomaly score.
For \textit{ETDddpm\text{-}E}, we integrate the results of both \textit{ETDddpm\text{-}R} and \textit{ETDddpm\text{-}F}, wherein an input is identified as an anomaly if either of the two methods detects it as such.
Note that we apply \textit{ETDddpm} for \textit{Electricity} but \textit{ETDddpm}$^+$ for \textit{Electricity-Theft} to reduce the impact of high-variance data. 
In \textit{LSTM\text{-}R}, we reimplement the model described in \cite{ref37} and add two FC layers as latent layers like \cite{ref10} to improve the performance. The model generates the reconstruction of an input sequence. We calculate the reconstruction error as the anomaly score.
In \textit{LSTM\text{-}F}, we reimplement the model described in \cite{ref38} to produce the forecasting sequence given an input sequence. We calculate the forecasting error as the anomaly score. 
In \textit{FC\text{-}R}, we reimplement the model described in \cite{ref9} to produce the reconstruction sequence given an input sequence. We calculate the reconstruction error as the anomaly score.
In \textit{VAE\text{-}R}, we reimplement the model described in \cite{ref10} to produce the reconstruction sequence given an input sequence. We calculate the reconstruction error as the anomaly score. In summary, the nomenclature convention employed here designates methods concluding with `-R' as REM, those ending with `-F' as FEM, and those concluding with `-E' as the Ensemble method.

\subsubsection{Experimental Results on \textit{Electricity}}
This section evaluates the proposed ETD methods on the four users of \textit{Electricity} with regular energy consumption. To better understand how normal users and dishonest users are distinguished, we illustrate two examples in Fig. \ref{fig:example-etd-electricity}. We employ a fixed reduction attack on the test dataset of User 3 to generate attack data. Then, we use \textit{ETDddpm} to produce reconstruction sequence, $\bm{\hat{x}}_{1:L}$, and forecasting sequence, $\bm{\hat{x}}_{L+1:L+T}$. In the upper left figure of Fig. \ref{fig:example-etd-electricity}, we calculate forecasting error as anomaly score, and the blue bins present the anomaly scores of normal sequences, and the orange bins present the anomaly scores of artificial sequences. Similarly, we calculate reconstruction errors as anomaly scores in the bottom left figure. From the figure, we can see that if the distance between the distributions of anomaly scores of normal and anomalous data is large, we can easily distinguish them. Besides, a larger distance leads to a larger AUC. 
Then, we employ all seven attacks mentioned in Section \ref{sec:attack methods} to generate anomalous data for each user and apply the proposed and baseline methods to detect those attacks. From Table \ref{tab:AUC scores user-specific}, we can see that REMs are usually better at detecting `fixed reduction', `partial reduction', and `random partial reduction' than FEM, which can be explained by the fact that reconstruction methods are sensitive to the input that is rarely met during training. On the contrary, FEMs perform better at `average consumption', `random average consumption', and `reverse', as the forecasting error \eqref{eq:error_F_adjust} places more emphasis on the shape of forecasting curves. 
From Table \ref{tab:AUC scores user-specific}, we can also conclude that for REMs, \textit{ETDddpm-R} shows better performance than \textit{LSTM-R}, \textit{FC-R}, and \textit{VAE-R} on \textit{Electricity}. For FEMs, \textit{ETDddpm-F} and \textit{LSTM-F} show similar performance on \textit{Electricity}. 

\begin{table*}[ht]
\renewcommand\arraystretch{1.}
    \caption{Average AUC Score and 5\%-TPR of Different ETD Methods on 278 users of \textit{Electricity}}
    \label{tab:scalability}
    \centering
    \begin{tabular}{c|c|c|c|c|c|c|c|c}
    \hline
    \multicolumn{9}{c}{\textbf{\textit{Average AUC Score for 278 Users}}}\\
    \hline
    
        {} &\makebox[0.024\textwidth]{\textbf{FR}} & \makebox[0.024\textwidth]{\textbf{PR}} & \makebox[0.024\textwidth]{\textbf{RPR}} & \makebox[0.024\textwidth]{\textbf{SBP}} &\makebox[0.024\textwidth]{\textbf{AC}} & \makebox[0.024\textwidth]{\textbf{RAC}} & \makebox[0.024\textwidth]{\textbf{REV}} & \makebox[0.024\textwidth]{\textbf{Ave}}\\
         \hline
         {\textbf{L-R}\cite{ref37}} &0.97 (0.30) & 0.97 (0.18) & 0.97 (0.23) & 0.94 (0.65) &0.99 (0.26) &0.99 (0.38) & 0.99 (0.68) &0.97 (0.38)   \\
         \hline
         {\textbf{L-F}\cite{ref38}} &0.98 (0.42) &0.98 (0.33) &0.98 (0.39) &0.94 (0.68) &0.98 (0.08) &1.00 (0.52) &0.99 (0.68) & \textbf{\textcolor{red}{0.98}} (\textbf{\textcolor{blue}{0.44}}) \\
         \hline
         {\textbf{VAE-R}\cite{ref10}} &0.98 (0.25) &0.97 (0.14) & 0.97 (0.22) & 0.94 (0.65) &0.98 (0.27) &1.00 (0.62) &0.99 (0.65) & \textbf{\textcolor{red}{0.98}} (0.40)  \\
         \hline
         {\textbf{ED-R (ours)}} &0.98 (0.26) &0.97 (0.12) &0.97 (0.18) &0.98 (0.64) &0.99 (0.22) &0.99 (0.41) &0.99 (0.62) &\textbf{\textcolor{red}{0.98}} (0.35)  \\
         \hline
         {\textbf{ED-F (ours)}} &0.78 (0.51) &0.85 (0.38) &0.94 (0.63) &1.00 (0.87) &0.98 (0.05) &0.99 (0.63) &1.00 (0.93) &{0.93} (\textbf{\textcolor{red}{0.57}})\\

         \hline

    \multicolumn{9}{c}{\textbf{\textit{Average 5\%-TPR for 278 Users}}}\\
    \hline
    
        {} &\makebox[0.024\textwidth]{\textbf{FR}} & \makebox[0.024\textwidth]{\textbf{PR}} & \makebox[0.024\textwidth]{\textbf{RPR}} & \makebox[0.024\textwidth]{\textbf{SBP}} &\makebox[0.024\textwidth]{\textbf{AC}} & \makebox[0.024\textwidth]{\textbf{RAC}} & \makebox[0.024\textwidth]{\textbf{REV}} & \makebox[0.024\textwidth]{\textbf{Ave}}\\
         \hline
         {\textbf{L-R}\cite{ref37}} &0.92 (0) & 0.91 (0) & 0.92 (0) & 0.87 (0.17) &0.96 (0) &0.98 (0) &0.98 (0.12) &0.93 (0.04)   \\
         \hline
         {\textbf{L-F}\cite{ref38}} &0.92 (0) &0.92 (0) &0.92 (0) &0.92 (0) &0.92 (0) &0.92 (0) &0.92 (0) & 0.92 (0)\\
         \hline
         {\textbf{VAE-R}\cite{ref10}} &0.93 (0) &0.92 (0)& 0.93 (0)& 0.87 (0.25) &0.96 (0) &0.98 (0.10) &0.98 (0.02) &0.94 (0.05)   \\
         \hline
         {\textbf{ED-R (ours)}} &0.93 (0) &0.92 (0) &0.93 (0) &0.95 (0.21) &0.97 (0) &0.98 (0.02) &0.98 (0.16) &\textbf{\textcolor{red}{0.95}} (0.06)\\
         \hline
         {\textbf{ED-F (ours)}} &0.05 (0) &0.29 (0) &0.49 (0) &0.96 (0.25) &0.95 (0) &0.96 (0.03) &0.98 (0.20) &{0.67} (\textbf{\textcolor{blue}{0.07}})\\
         \hline
         {\textbf{ED-E (ours)}} &0.95 (0.58) &0.95 (0.53) &0.95 (0.51) &0.95 (0.48) &0.95 (0.51) &0.95 (0.49) &0.95 (0.40) &\textbf{\textcolor{red}{0.95}} (\textbf{\textcolor{red}{0.50}})\\
         \hline
    \end{tabular}
    \begin{tablenotes}
        \footnotesize
        \item Abbreviations are the same as Table \ref{tab:AUC scores user-specific} and Table \ref{tab:AUC scores of single models and the ensemble model}; numbers in brackets denote the lowest performance among 278 users.
    \end{tablenotes}
\end{table*}
\subsubsection{Experimental Results on \textit{Electricity-Theft}} \label{sec:Experiment results on Electricity-Theft}
In this section, we evaluate the proposed methods on four users of \textit{Electricity-Theft}, in which two users have a high-variance energy consumption, one user has a medium-variance consumption, and one user has a low-variance energy consumption. For this dataset, we apply \textit{ETDddpm}$^+$ for \textit{ETDddpm-F}, \textit{ETDddpm-R}, and \textit{ETDddpm-E}, instead of the original \textit{ETDddpm} to mitigate the impact of the high variance. As shown in Table \ref{tab:AUC scores user-specific}, all baseline methods cannot work for high-variance data, which shows that they can neither learn the pattern of high-variance data nor the relationship between the three attributes. On the contrary, \textit{ETDddpm}-based methods can work well on these high-variance smart grid data. For the user exhibiting medium-variance energy consumption, i.e., User 4, most baseline methods present a poor performance while \textit{ETDddpm}-based methods can work perfectly. For the user exhibiting low-variance energy consumption, i.e., User 3, most methods effectively detect instances of energy theft.

\subsubsection{Enhancement Brought by the Ensemble Method} \label{Section:results of ensemble method}
Although the AUC scores seem satisfactory in Table \ref{tab:AUC scores user-specific}, a single REM or FEM may be insufficient to detect all energy thefts for a given user. 
For example, \textit{ETDddpm\text{-}F} works well for User 1 of \textit{Electricity} but cannot identify some attacks on User 3 of \textit{Electricity}. On the other hand, \textit{ETDddpm\text{-}R} works well for User 3 of \textit{Electricity} but cannot identify some attacks on User 1 of \textit{Electricity}. To construct an effective ETD method for all users, we propose the ensemble approach, \textit{ETDddpm\text{-}E}.
We assess \textit{ETDddpm\text{-}E} on User 1 and User 3 of \textit{Electricity} to show how the ensemble method can improve the ETD performance. 
First, we set the maximum \textit{FPR}, $\alpha$, to be 5\% for each single REM and FEM. For the ensemble model \textit{ETDddpm\text{-}E}, we set the maximum \textit{FPR} to be 2.5\% for each submodule, i.e., \textit{ETDddpm\text{-}R} and \textit{ETDddpm\text{-}F}, in order to achieve the same maximum \textit{FPR} as other methods, i.e., 5\%. 
In Table \ref{tab:AUC scores of single models and the ensemble model}, it is evident that, under a 5\% maximum False Positive Rate (FPR), most baseline methods exhibit poor performance, whereas \textit{ETDddpm\text{-}E} demonstrates the highest True Positive Rate (TPR) with a significant lead over the second-highest TPR. Similarly, under a 10\% maximum FPR, \textit{ETDddpm\text{-}E} maintains a substantial lead on User 1 and secures the second position on User 3 with a slight margin.
Summarizing the results from Table \ref{tab:AUC scores of single models and the ensemble model}, it is clear that \textit{ETDddpm\text{-}E} consistently achieves a high TPR for both users, even at a low FPR (5\%), thereby enhancing performance compared to the individual models \textit{ETDddpm\text{-}F} and \textit{ETDddpm\text{-}R}. For the remaining users in the \textit{Electricity} and \textit{Electricity-Theft} datasets, \textit{ETDddpm\text{-}E} attains a 100\% TPR at a small FPR, as either \textit{ETDddpm\text{-}F} or \textit{ETDddpm\text{-}R} consistently achieves an AUC score of 1.00 for a certain attack. Due to space constraints, we omit the experimental results on those users in Table \ref{tab:AUC scores of single models and the ensemble model}.

\subsubsection{Scalability} \label{sec:scalability}
It should be noted that in the above experiments, we have considered user-specific scenarios. Now,  in this section, we show how our proposed scheme works on multiple-user scenarios. In this regard, we choose 278 users from \textit{Electricity} excluding users with missing values and concatenate the data of 278 users so the input space $\mathcal{X} = \mathbb{R}^1$ for user-specific scenarios is changed to $\mathcal{X} = \mathbb{R}^{278}$. 
Only one user is under attack each time. We repeat the experiment 278 times to attack all the users and calculate the average ETD performance in Table \ref{tab:scalability}. We can see that all baseline methods achieve a good performance w.r.t. the average AUC score. However, they are \textbf{ineffective} for some users as shown with the lowest performance in the brackets. The average of the lowest AUC is around 0.5, which means single REM and FEM cannot work for the stealthiest attacks. As a result, with a maximum of 5\% FPR, the TPR for the stealthiest attacks is very low. On the contrary, the ensemble method achieves a significant improvement for the stealthiest users w.r.t. 5\%-TPR.
\vspace{-5mm}
\section{Conclusion} \label{sec:5} 
In this paper, we delineate the inherent constraints of existing unsupervised ETD methods. Specifically, we observe that current REMs for ETD encounter challenges in consistently achieving high performance across diverse user profiles. These methods also exhibit limitations in accurately identifying instances of energy theft within high-variance smart grid data.
To address these issues, we propose a DDPM-based ensemble ETD method, denoted as \textit{ETDddpm\text{-}E}. This method integrates the principles of REM and FEM and leverages DDPM to generate reconstruction and forecasting sequences. From experimental results, We observe that \textit{ETDddpm\text{-}R} and \textit{ETDddpm\text{-}F} demonstrate distinct performances under various attacks and user scenarios. In general, these two types of methods are complementary to each other, and their complementary nature is harnessed in \textit{ETDddpm\text{-}E}, resulting in consistently high performance across all types of attacks and users for both user-specific and multiple-user scenarios.
Furthermore, our analysis indicates that \textit{ETDddpm\text{-}E} achieves nearly perfect performance for all four users in \textit{Electricity-Theft}, whereas baseline methods cannot work for high-variance users.

\bibliographystyle{IEEEtran}
\bibliography{reference}

\end{document}